\documentclass[letterpaper, 10 pt, journal, twoside]{IEEEtran}
\usepackage{graphicx} 
\usepackage{amsmath} 
\usepackage{amssymb} 
\usepackage[T1]{fontenc}
\usepackage{multirow}
\usepackage{makecell}
\usepackage{capt-of}
\usepackage{adjustbox}
\usepackage{threeparttable}
\usepackage{booktabs}
\usepackage{pifont}
\usepackage{float}
\usepackage{algorithm}
\usepackage{algpseudocode}
\usepackage{array,stfloats}

\usepackage{xcolor}
\definecolor{linkcolor}{HTML}{ED1C24}
\usepackage[pagebackref=false,breaklinks=true,colorlinks,urlcolor=blue,citecolor=blue,linkcolor=linkcolor,bookmarks=false, ]{hyperref}
\newcommand{\cmark}{\ding{51}}

\newcommand{\methodname}{UOIS-SAM}
\newcommand{\hdnetname}{OLQA}

\author{Rui Cao\textsuperscript{1},
Mingyuan He\textsuperscript{1},
Yunfan Yu\textsuperscript{1},
Jia-Hui Pan\textsuperscript{2},
Biqi Yang\textsuperscript{3},
Wei Chen\textsuperscript{1},
Xieyuanli Chen\textsuperscript{1,4},
Jiangliu Wang\textsuperscript{1,5},
and Yun-Hui Liu\textsuperscript{1}\textsuperscript{\ding{41}}%
\thanks{This work is supported in part by Research Grants Council (RGC) of Hong Kong under Grant 14207324, in part by the HK RGC AoE under AoE/E-407/24-N, in part by the InnoHK initiative of the Innovation and Technology Commission of the Hong Kong Special Administrative Region Government via the Hong Kong Centre for Logistics Robotics, and in part by the CUHK T Stone Robotics Institute.
\emph{(\ding{41} Corresponding author: Yun-Hui Liu.)}}%
\thanks{\textsuperscript{1} Department of Mechanical and Automation Engineering, The Chinese University of Hong Kong, Hong Kong, China.
E-mails: rcao@mae.cuhk.edu.hk (Rui Cao); yhliu@mae.cuhk.edu.hk (Yun-Hui Liu).}%
\thanks{\textsuperscript{2} Department of Computer Science and Engineering, The Chinese University of Hong Kong, Hong Kong, China.}%
\thanks{\textsuperscript{3} China Academy of Space Technology, Beijing, China.}%
\thanks{\textsuperscript{4} National University of Defense Technology, Hunan, China.}%
\thanks{\textsuperscript{5} Sino-German College of Intelligent Manufacturing, Shenzhen Technology University, Shenzhen, China.}%
}

\newcommand{\rui}[1]{{#1}}

\begin{document}

\title{\methodname: Adapting Promptable Foundation Model for Zero-shot Object Instance Segmentation
}

\maketitle

\begin{abstract}
Zero-shot Object Instance Segmentation is crucial for autonomous robots operating in unstructured environments.
Previous approaches often rely on full supervision from large-scale synthetic tabletop datasets for task-specific training.
In this paper, we propose \methodname, a data-efficient framework for zero-shot object instance segmentation that adapts frozen promptable segmentation models using limited synthetic supervision.
\methodname\ links three stages: (i) a Heatmap-based Prompt Generator to generate point prompts on foreground objects, (ii) an Object-Level Quality Adapter to estimate candidate quality for individual visible objects, and (iii) Global Candidate Selection and Integration to consolidate masks into a coherent instance set using these estimates, spatial overlap, and containment.
We conduct extensive experiments on OCID, OSD, and the photometrically challenging HouseCat6D dataset. Using only 10\% of Tabletop Object Dataset (TOD) training scenes for task adaptation, \methodname\ achieves the highest Overlap and Boundary F-measures among compared methods across all three benchmarks, demonstrating strong data efficiency and robustness in tabletop clutter.
Downstream robotic grasping evaluations further show improved Target AP@10 on SuctionNet-1Billion and higher execution success rates in real-world robotic experiments, including cluttered scenes with transparent and reflective objects.
\emph{We will release our code, model, and data for both training and evaluation upon publication.}
\end{abstract}


\section{Introduction}
\rui{Zero-shot object instance segmentation aims to segment individual objects in unstructured scenes without predefined categories or target-domain training.} It serves as a key foundation for downstream robotic tasks such as 6-DoF pose estimation~\cite{nguyen2024gigapose, labbe2022megapose,li2022sim}, grasp synthesis~\cite{mousavian20196,wen2022transgrasp, cao2024uncertainty}, and robotic manipulation~\cite{manuelli2019kpam,2024EquivAct,chen2023learning}, \rui{which require accurate masks to distinguish individual physical objects from their surroundings. In robotics, this problem is commonly studied as Unseen Object Instance Segmentation (UOIS).}

\begin{figure}[!t]
\includegraphics[width=\columnwidth]{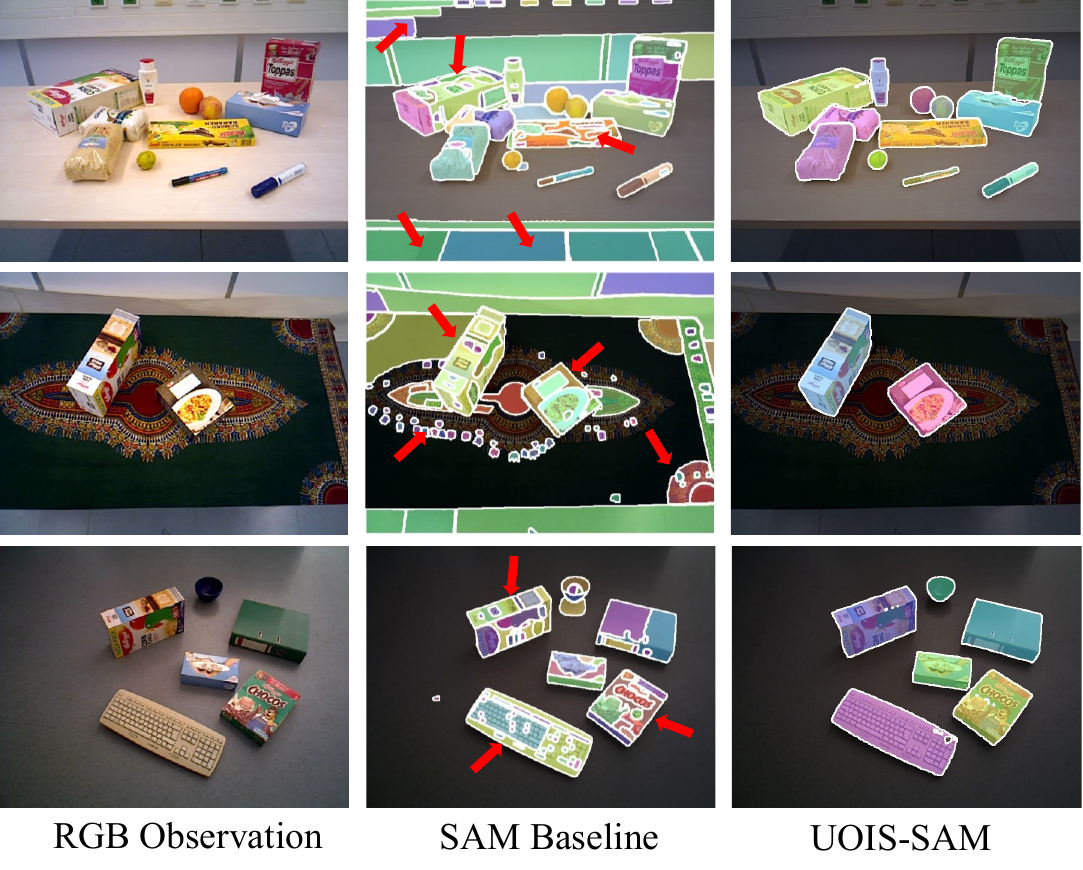}
\centering
\caption{Comparison results of our \methodname\ and the SAM baseline. The red arrows highlight common issues of the SAM baseline, such as background mis-segmentation and severe over-segmentation. \methodname\ demonstrates notably fewer background segmentation errors and generates masks of higher accuracy, particularly for texture-rich objects.}
\label{fig:teaser}
\end{figure}

\rui{Many task-specific UOIS methods rely on heavy synthetic supervision to learn objectness and grouping via depth-based seeding, feature clustering, or occlusion reasoning~\cite{xie2021unseen,xiang2020learning,back2022unseen,lu2022mean}. However, they remain vulnerable to the sim-to-real gap, especially on transparent and reflective objects. Promptable segmentation foundation models---such as SAM~\cite{kirillov2023segment}, Semantic-SAM~\cite{li2024segment}, and SEEM~\cite{zou2023segment}---offer an appealing alternative: pretrained on massive, diverse imagery, they already possess strong, generalizable mask-generation capabilities. Yet applying them directly to robotic UOIS still yields severe background false positives, over-segmentation, and duplicate masks (Fig.~\ref{fig:teaser}).}

\rui{Our key insight is that frozen promptable foundation models can already produce accurate object-level masks. Rather than relearning how to generate masks, the core problem is identifying true object instances among the generated candidates, which breaks down into three key subproblems. First, the framework must determine \emph{where to query}, because uniform prompts frequently fall on background regions. Second, it must determine \emph{which candidate to trust} within each prompt, because a single point can yield plausible part, object, or multi-object masks whose native scores may not favor the individual object. Third, it must determine \emph{which candidates to retain} across prompts, because independently decoded masks often overlap or duplicate the same instance.}

\rui{Guided by this principle, we propose \methodname, a lightweight framework that solves these three subproblems around a frozen promptable foundation model through three sequential stages. To determine where to query, the Heatmap-based Prompt Generator (HPG) uses visual and relative-depth features from a frozen monocular depth model to predict a foreground mask and an object-center heatmap, generating foreground point prompts while guiding candidate filtering. To decide which candidate to trust, the Object-Level Quality Adapter (\hdnetname) learns lightweight residual corrections to each model's native scores using visible-object IoU supervision. This mitigates within-prompt granularity ambiguity without altering mask geometry and provides adapted object quality estimates downstream. To choose which candidates to retain, Global Candidate Selection and Integration (GCSI) combines HPG's foreground prior, \hdnetname's adapted scores, and pairwise overlap and containment penalties to suppress redundancy and construct a coherent, non-overlapping instance set. Because the pretrained backbones remain frozen and only the lightweight HPG layers and \hdnetname\ are trained on limited synthetic data, the framework is data-efficient and applicable across promptable models.}

\rui{We evaluate \methodname\ on OCID, OSD, and HouseCat6D using only 10\% of the TOD training scenes for task adaptation. With prompts and frozen SAM candidates fixed, \hdnetname\ recovers 85.8\%--86.6\% of the Mean Selected IoU gap between native scoring and oracle selection (Tab.~\ref{tab:candidate_scoring}). Under the evaluated adaptation settings, adapting candidate quality scores while keeping the mask generator frozen outperforms direct mask decoder fine-tuning using less than 5\% of the trainable parameters (Tab.~\ref{tab:candidate_adaptation}). The same quality objective consistently improves segmentation across SAM, Semantic-SAM, and SEEM on all three datasets. Finally, downstream evaluations show higher Target AP@10 on SuctionNet-1Billion and higher suction success rates in real tabletop clutter with transparent and reflective objects.}

\begin{figure*}[t]
\centering
\includegraphics[width=\textwidth]{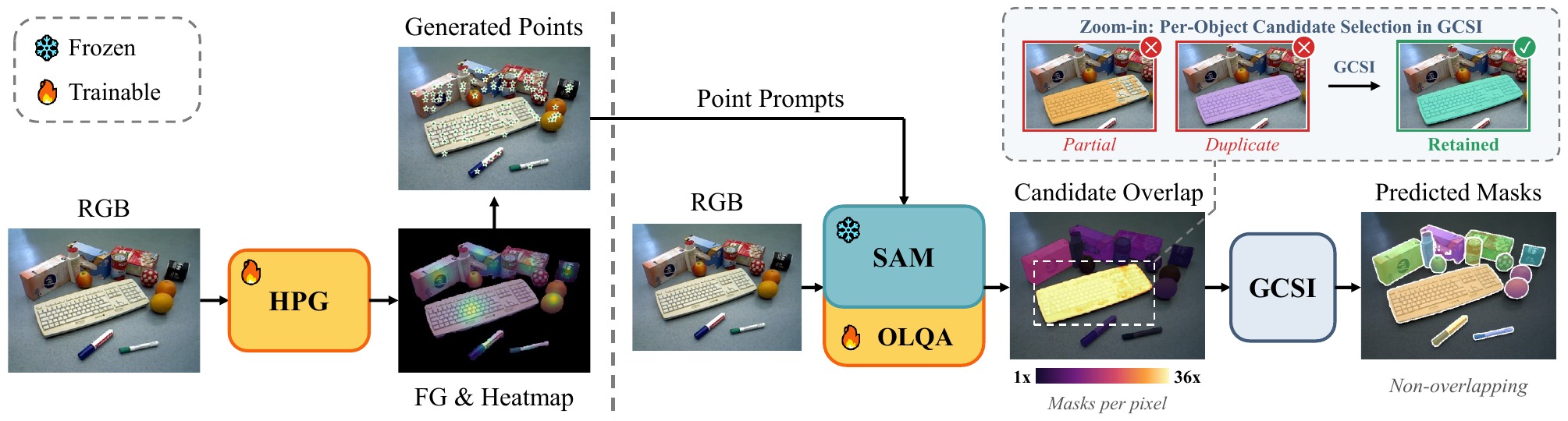}
\caption{Overview of \methodname. Given an RGB input, a Heatmap-based Prompt Generator (HPG) \rui{predicts a foreground mask (FG) and object-center heatmap to generate} point prompts for SAM. An Object-Level Quality Adapter (\hdnetname) \rui{attached to SAM} refines \rui{candidate quality} scores \rui{within each prompt, while GCSI resolves heavy multi-prompt candidate overlap (up to $36\times$ masks per pixel) by suppressing partial and duplicate proposals (top-right zoom-in) into a non-overlapping instance set.} The \rui{task-specific HPG layers} and \hdnetname\ are trainable; SAM remains frozen.}
\label{fig:overview}
\end{figure*}

The main contributions of this work are summarized as follows:
\begin{itemize}
    \item \rui{We formulate the adaptation of frozen promptable models for robotic UOIS as three key subproblems, and instantiate this principle in \methodname\ by linking foreground prompting (HPG), object-level quality adaptation (\hdnetname), and overlap-aware consolidation (GCSI).}
    \item We show that frozen promptable models can produce accurate object candidates that are overlooked by native scores, and introduce \hdnetname\ to learn quality scores for individual visible objects across SAM, Semantic-SAM, and SEEM.
    \item \rui{Using only 10\% of synthetic TOD scenes for task adaptation, \methodname\ achieves the highest Overlap and Boundary F-measures among compared methods on OCID, OSD, and HouseCat6D, while improving target-aware suction grasping on SuctionNet-1Billion and in real-world clutter with transparent and reflective objects.}
\end{itemize}

\section{Related Work}
\subsection{\rui{Unseen Object Instance Segmentation}}
Traditional instance segmentation focuses on detecting object instances within predefined classes~\cite{he2017mask, wang2020solo,cheng2022masked}.
To reduce the need for manual annotations, unsupervised class-agnostic instance segmentation methods, such as FreeSOLO~\cite{wang2022freesolo} and CutLER~\cite{wang2023cut}, learn to discover and segment objects without manual instance labels. More recently, promptable segmentation models, including SAM~\cite{kirillov2023segment}, Semantic-SAM~\cite{li2024segment}, and SEEM~\cite{zou2023segment}, have demonstrated general mask prediction capabilities using point and other prompts. However, generating masks from prompts does not necessarily produce a complete and non-redundant set of object instances in cluttered scenes.

In robotics, Unseen Object Instance Segmentation (UOIS) aims to segment arbitrary objects without restricting predictions to predefined task-specific categories. Existing UOIS methods often learn class-agnostic objectness using synthetic training data~\cite{xie2020best,back2022unseen}.
Earlier works explored two-stage networks to process RGB and depth images separately~\cite{xie2020best,xie2021unseen}, while Durner~\textit{et al.}~\cite{durner2021unknown} utilized stereo image pairs to predict instance masks in cluttered environments.
Unseen Clustering Network (UCN)~\cite{xiang2020learning} proposed a method that extracts RGB and depth features, grouping them at the pixel level using mean shift clustering to distinguish unseen objects.
Building on this, MSMFormer~\cite{lu2022mean} replaced the mean shift clustering module with differentiable transformer layers~\cite{vaswani2017attention}.
UOAIS-Net~\cite{back2022unseen} introduces hierarchical amodal perception, which is particularly beneficial for cluttered environments, while Yang~\textit{et al.}~\cite{yang2023improving} emphasize boundary awareness by using a dedicated head to incrementally refine boundary quality.
These methods generally rely on task-specific training for object instance segmentation in robotic scenes.

Pretrained segmentation models have also been applied to robotic object perception. SAM-6D~\cite{lin2024sam} uses SAM-generated object proposals for zero-shot 6D object pose estimation. More closely related to robotic UOIS, ZISVFM~\cite{zisvfm} introduces a training-free approach that generates SAM proposals from colorized depth images, filters non-object masks using pretrained visual features, and uses point prompts obtained by K-Medoids clustering for further segmentation. \rui{In contrast to data-intensive RGB-D methods and depth-dependent SAM pipelines such as ZISVFM, \methodname\ operates from RGB input alone by keeping pretrained promptable models frozen and learning lightweight foreground prompting and object-level quality adaptation from limited synthetic supervision.}

\subsection{\rui{Segmentation Quality and Instance Selection}}
\rui{Mask Scoring R-CNN~\cite{huang2019mask} predicts mask IoU to better align instance scores with segmentation quality. In robotic UOIS, RICE~\cite{xie2022rice} learns to score segmentation graphs and evaluates alternatives generated by splitting, merging, adding, and deleting masks. UncOS~\cite{fang2024embodied} generates multiple object segmentation hypotheses from repeated queries to pretrained models and estimates confidence from the consistency of these hypotheses.}

\rui{Our framework instead estimates the quality of individual candidates produced by frozen promptable segmentation models. Using synthetic instance supervision, \hdnetname\ learns scores based on each candidate's IoU with the visible object containing the prompt. The adapted scores guide mask selection within prompts and candidate retention across prompts, while GCSI combines these estimates with spatial overlap and containment to construct the final instance set.}

\section{Method}
\label{sec:method}
As illustrated in Fig.~\ref{fig:overview}, given an \textbf{RGB} observation of a cluttered scene, our objective is to segment all \rui{visible object instances without predefined categories.} \rui{We train HPG and \hdnetname\ on the synthetic Tabletop Object Dataset (TOD)~\cite{xie2021unseen}. Here, zero-shot evaluation refers to applying the resulting framework to the target datasets without further training or adaptation.}

\noindent\textbf{\rui{Promptable Segmentation Model Baseline.}}
\rui{Pretrained promptable segmentation models, such as SAM, Semantic-SAM, and SEEM, generate candidate masks from point prompts. A straightforward approach to automatic instance segmentation is to sample point prompts on a regular grid and select candidate masks using the models' native scores. For example, SAM's automatic mask generator uses a regular grid of point prompts (e.g., $32\times32$) to generate mask proposals.}
However, as shown in \rui{Fig.~\ref{fig:teaser}}, this simple approach presents several challenges.
\rui{First, uniformly sampled prompts may fall on background regions, producing false-positive masks. Second, native candidate scores do not necessarily favor masks at the desired object-level granularity. Finally, masks generated from different prompts may overlap or represent the same object, requiring additional instance selection.}

\noindent\textbf{Proposed Framework.}
To address these challenges, we propose \methodname, \rui{a lightweight framework that adapts pretrained promptable segmentation models to zero-shot object instance segmentation in robotic scenes.}
\rui{Our framework consists of three stages.}
The first is the Heatmap-based Prompt Generator (HPG), which generates point prompts by predicting a class-agnostic heatmap and performing foreground-background segmentation \rui{to place prompts within foreground regions}.
\rui{These prompts are fed into a frozen segmentation model to generate candidate masks.}
\rui{The second stage is the Object-Level Quality Adapter (OLQA), which uses a lightweight residual multilayer perceptron (MLP) to refine native candidate scores using features extracted from each model. These scores guide candidate selection and filtering to favor masks representing individual objects.}
\rui{The third stage, Global Candidate Selection and Integration (GCSI), selects the final instance set by accounting for overlap and containment among candidates across prompts.}
\rui{The pretrained segmentation models remain frozen, while HPG and OLQA are trained with limited synthetic supervision.}
The detailed designs \rui{are presented below}.

\subsection{Heatmap-based Prompt Generator}
\label{sec:HPG}
\rui{HPG predicts object locations and foreground regions to generate point prompts automatically.} Given an RGB observation $\mathbf{I} \in \mathbb{R}^{H\times W\times 3}$ of a scene, HPG regresses a heatmap \rui{$\hat{\mathbf{H}} \in \mathbb{R}^{H\times W}$ that indicates representative points of class-agnostic objects. To construct the ground truth (GT) heatmap, we use image moments~\cite{gonzales1987digital} to calculate the 2D centroid of the largest visible connected component of each instance. If the centroid lies outside the component, we use the interior point with the largest distance to the object boundary.}

\rui{We construct the GT heatmap $\mathbf{H}\in[0,1]^{H\times W}$ from the instance keypoints. Let $p=(x,y)$ denote a pixel, $\boldsymbol{\mu}_j$ the keypoint of instance $j$, and $\sigma_j$ the Gaussian scale determined by its bounding box~\cite{zhou2019objects}, with $\sigma_j\geq 2$. For $J$ GT instances, the heatmap is the maximum of their Gaussian responses:}
\begin{equation}
\rui{\mathbf{H}_p=\max_{1\leq j\leq J}\exp\!\left(-\frac{\|p-\boldsymbol{\mu}_j\|_2^2}{2\sigma_j^2}\right).}
\end{equation}
\rui{We supervise the heatmap using Mean Squared Error (MSE) over the image domain $\Omega$, with $N_{\mathrm{pix}}=|\Omega|=HW$:} 
\begin{equation}
\rui{\mathcal{L}_h = \frac{1}{N_{\mathrm{pix}}} \sum_{p\in\Omega} \left( \mathbf{H}_p - \hat{\mathbf{H}}_p \right)^2,}
\end{equation}
\rui{Here, $\hat{\mathbf{H}}_p$ denotes the predicted heatmap value at pixel $p$.}

Meanwhile, to prevent point prompts from scattering in background regions, we incorporate a foreground segmentation network. To achieve this, we formulate foreground segmentation as a pixel-wise classification problem. 
Specifically, HPG produces a \rui{foreground-background logit map $\hat{\mathbf{F}} \in \mathbb{R}^{H\times W\times 2}$}.
\rui{The GT label map $\mathbf{F}\in\{0,1\}^{H\times W}$ is obtained by merging the GT instance masks, with $\mathbf{F}_p=1$ for object pixels.} To address the issue of data imbalance, we optimize this objective using a weighted cross-entropy loss~\cite{xie2021unseen}. The loss function is defined as:
\begin{equation}
\rui{\mathcal{L}_{fg} = \sum_{p\in\Omega} w_p \, \mathrm{CE}(\hat{\mathbf{F}}_p, \mathbf{F}_p),}
\end{equation}
\rui{Here, $\hat{\mathbf{F}}_p\in\mathbb{R}^2$ denotes the foreground logits, $\mathbf{F}_p$ is the GT label, and $\mathrm{CE}$ is the cross-entropy loss. Each weight $w_p$ is inversely proportional to the frequency of its GT class, with weights normalized to sum to one.} The total loss used to optimize HPG is given by:
\begin{equation}
\mathcal{L}_{\mathrm{HPG}} = \lambda_1 \mathcal{L}_{fg} + \lambda_2 \mathcal{L}_{h},
\end{equation}
where \rui{$\lambda_1$ and $\lambda_2$ balance foreground segmentation and heatmap regression}.

\rui{We use a frozen Depth Anything V2 (DAV2)~\cite{depth_anything_v2} as the backbone of HPG. Given an RGB image, DAV2 provides both multi-scale image features and a relative-depth prediction. We aggregate the image features with a DPT~\cite{ranftl2021vision} decoder and upsample them to the image resolution. To incorporate geometric cues, we normalize the relative-depth map within each image, project it into a lightweight depth feature, and concatenate it with the upsampled image feature. The concatenated representation is further fused by convolutional layers and used to predict foreground logits and a heatmap of object locations, $\rui{\operatorname{Concat}(\hat{\mathbf{F}},\hat{\mathbf{H}}) \in \mathbb{R}^{H\times W\times 3}}$. We then generate point prompts from the heatmap within the predicted foreground regions. We adopt DAV2 with relative depth to improve object discovery for photometrically challenging objects, as demonstrated in Sec.~\ref{sec:ablations}.}

\begin{figure}[t]
\includegraphics[width=\columnwidth]{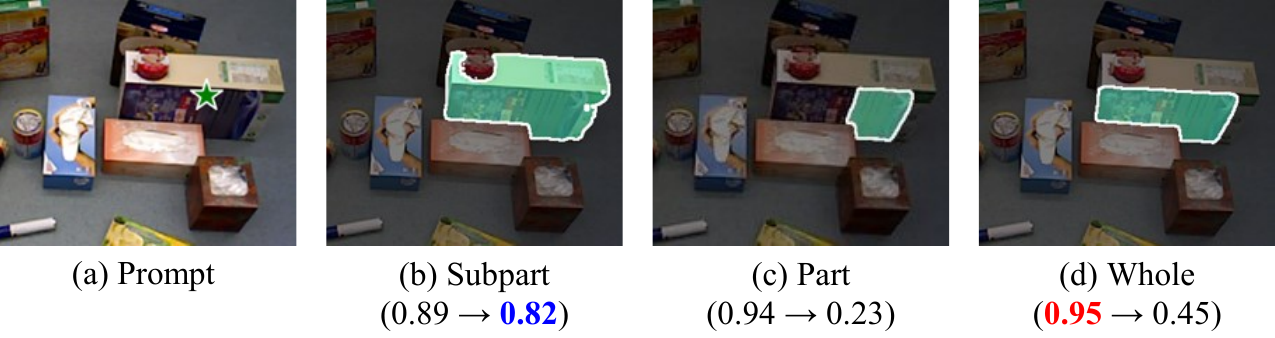}
\caption{\rui{Example of candidate score adaptation using OLQA with SAM on an OCID~\cite{suchi2019easylabel} image. Given a point prompt (denoted by $\textcolor[RGB]{0, 255, 0}{\star}$), SAM produces three candidate masks with native IoU scores. The native scores favor an incorrect mask (\textcolor{red}{red}), while OLQA selects the candidate better matching the GT object (\textcolor{blue}{blue}).}}
\label{fig:olqa_example}
\end{figure}

\subsection{Object-Level Quality Adapter}
\label{sec:olqa}
\rui{Promptable segmentation models can produce multiple masks for a single point, representing an object part, an individual object, or a larger region containing neighboring objects. Robotic UOIS requires masks that correspond to individual visible objects. Our analysis with identical prompts and frozen SAM masks shows that the best available candidates often have substantially higher IoU with the target objects than the masks selected using SAM's original scores (Tab.~\ref{tab:candidate_scoring}). Thus, suitable masks may already be available even when the original scores fail to select them. Motivated by this observation, we introduce the Object-Level Quality Adapter (OLQA) to learn quality scores that favor masks corresponding to individual objects, without changing the pretrained mask predictions.}

\rui{Given a point prompt $p_i$, a frozen segmentation model produces $K_i$ candidate masks $\{\hat{\mathbf{M}}_{ik}\}_{k=1}^{K_i}$, with native scores $s_{ik}$ and feature vectors $\mathbf{z}_{ik}$ constructed from the candidate and prompt representations. OLQA employs a lightweight residual MLP to predict adapted quality scores:}
\begin{equation}
\rui{
\hat{s}_{ik}
= s_{ik} + f_{\theta}(\mathbf{z}_{ik}),
}
\label{eq:olqa}
\end{equation}
\rui{where $f_{\theta}$ predicts a scalar score correction for each candidate. The adapted scores guide candidate selection within each prompt and candidate retention during subsequent instance consolidation. Only the adapter parameters $\theta$ are optimized, while the pretrained segmentation model remains frozen.}

\rui{To align candidate scores with object-level segmentation, we supervise each candidate using its IoU with the GT visible instance containing the prompt. Background prompts are assigned zero quality. Specifically, the target score is}
\begin{equation}
\rui{
s_{ik}^{*} =
\begin{cases}
\operatorname{IoU}(\hat{\mathbf{M}}_{ik}, \mathbf{G}(p_i)),
& p_i \in \mathcal{F},\\
0, & p_i \notin \mathcal{F},
\end{cases}
}
\label{eq:olqa_target}
\end{equation}
\rui{where $\mathbf{G}(p_i)$ denotes the GT visible instance containing $p_i$, and $\mathcal{F}=\{p\in\Omega:\mathbf{F}_p=1\}$ is the GT object foreground. We optimize OLQA using the MSE loss:}
\begin{equation}
\rui{
\mathcal{L}_{\mathrm{OLQA}}
= \frac{1}{N_{\mathrm{cand}}}
\sum_{i=1}^{M}\sum_{k=1}^{K_i}
\left(\hat{s}_{ik}-s_{ik}^{*}\right)^2,
}
\label{eq:olqa_loss}
\end{equation}
\rui{where $M$ is the number of prompts and $N_{\mathrm{cand}}=\sum_{i=1}^{M}K_i$ is the total number of candidates.}

\noindent\textbf{\rui{Adaptation to Different Segmentation Models.}}
\rui{We instantiate OLQA for SAM, Semantic-SAM, and SEEM using the features available from each model. For SAM, we concatenate each mask token with the shared IoU token and refine the native IoU prediction. For Semantic-SAM, we use decoder features of individual candidates to refine their native IoU predictions. For SEEM, we use mask query and spatial prompt embeddings, together with their matching scores, to adapt the native matching scores. Each model uses an independently trained residual MLP with the same object-level supervision, without changing the predicted masks. Cross-architecture adapter designs, feature taps, input dimensionalities, and parameter counts are detailed in Appendix~\ref{app:olqa}.}

\subsection{\rui{Global Candidate Selection and Integration}}
\label{sec:gcsi}
\rui{Candidate masks generated from different prompts may overlap or represent the same object. GCSI uses candidate scores and spatial overlap to select the final set of instances.
For each prompt $p_i$, we retain the candidate $k_i^*=\arg\max_k\hat{s}_{ik}$, with binary mask $\mathbf{B}_i=\hat{\mathbf{M}}_{ik_i^*}$ and adapted score $q_i=\hat{s}_{ik_i^*}$.
We measure overlap and containment between retained candidates and greedily accept masks according to their scores and redundancy with previously selected masks.
Finally, the selected masks are integrated into a non-overlapping instance label map.
Algorithm~\ref{alg:gcsi} summarizes this procedure.}

\begin{algorithm}[t]
\caption{\rui{Global Candidate Selection and Integration}}
\label{alg:gcsi}
\begin{algorithmic}[1]
\Require Candidate masks and adapted quality scores for each prompt
\Ensure Instance label map $\mathbf{L}$ with no overlapping labels

\State Retain the highest-scoring candidate from each prompt
\State Filter candidates by quality, foreground agreement, and mask area
\State Compute pairwise redundancy $R_{ij}$ using
Eqs.~\eqref{eq:gcsi_relation} and~\eqref{eq:gcsi_penalty}
\State Sort candidates in descending order of $q_i$
\State Initialize selected set $\mathcal{S}\gets\emptyset$

\For{each candidate $i$ in the sorted order}
    \If{$\exists j\in\mathcal{S}:
    \mathrm{IoU}_{ij}\geq\tau_u^{\mathrm{hard}}
    \lor \mathrm{IoS}_{ij}\geq\tau_s^{\mathrm{hard}}$}
        \State \textbf{continue}
    \EndIf
    \State Compute $\Delta_i$ using Eq.~\eqref{eq:gcsi_marginal}
    \If{$\Delta_i \ge \tau_\Delta$}
        \State $\mathcal{S}\gets\mathcal{S}\cup\{i\}$
    \EndIf
\EndFor

\State Sort $\mathcal{S}$ in descending order of $\Delta_i$
\State Initialize $\mathbf{L}$ as background
\For{each $i\in\mathcal{S}$}
    \State Compute visible area and visible fraction of $\mathbf{B}_i$
    \If{either final visibility constraint is violated}
        \State \textbf{continue}
    \EndIf
    \State Assign the unoccupied pixels of $\mathbf{B}_i$ to a new instance in $\mathbf{L}$
\EndFor
\State \Return $\mathbf{L}$
\end{algorithmic}
\end{algorithm}

\rui{Let $a_i=|\mathbf{B}_i|$ denote the area of retained mask $i$, and $I_{ij}=|\mathbf{B}_i\cap\mathbf{B}_j|$ the intersection area between masks $i$ and $j$. We measure their overlap as}
\begin{equation}
\label{eq:gcsi_relation}
\rui{
\begin{aligned}
\mathrm{IoU}_{ij} &= \frac{I_{ij}}{a_i+a_j-I_{ij}},\\
\mathrm{IoS}_{ij} &= \frac{I_{ij}}{\min(a_i,a_j)},\\
c_{ij} &= \frac{I_{ij}}{\max(a_i,a_j)}.
\end{aligned}
}
\end{equation}
\rui{Here, $\mathrm{IoU}_{ij}$ measures overall overlap, $\mathrm{IoS}_{ij}$ measures the fraction of the smaller mask covered by the intersection, and $c_{ij}$ measures
the fraction of the larger mask covered by the intersection.
We combine these relations into the pairwise redundancy penalty}
\begin{equation}
\label{eq:gcsi_penalty}
R_{ij} =
\rui{\lambda_u\phi(\mathrm{IoU}_{ij};\tau_u)
+\lambda_s\phi(\mathrm{IoS}_{ij};\tau_s)
+\lambda_c c_{ij}.}
\end{equation}
\rui{Here, $\phi(x;\tau)=\max(x-\tau,0)/(1-\tau)$ activates a penalty only when the corresponding overlap exceeds the soft threshold $\tau$. The thresholds $\tau_u$ and $\tau_s$ suppress penalties from incidental overlap, while $\lambda_u$, $\lambda_s$, and $\lambda_c$ balance the three redundancy terms.}

\rui{Given the current selected set $\mathcal{S}$, candidate $i$ contributes its own quality $q_i$ while introducing redundancy with candidates already in $\mathcal{S}$. We quantify its effective contribution by the marginal utility}
\begin{equation}
\label{eq:gcsi_marginal}
\rui{
\Delta_i=q_i-\sum_{j\in\mathcal{S}}R_{ij}.
}
\end{equation}
\rui{A distinct candidate therefore largely preserves its quality score, whereas a candidate that duplicates existing instances receives a lower $\Delta_i$. GCSI greedily accepts candidates whose marginal utility exceeds $\tau_\Delta$. The selected masks are then integrated in descending order of $\Delta_i$, preserving previously assigned pixels to produce a non-overlapping instance label map. Further implementation details are provided in Sec.~\ref{sec:exp}.}

\section{Experiments}
\subsection{Implementation Details}
\label{sec:exp}
\noindent\textbf{Benchmark.} Following prior works~\cite{xiang2020learning,back2022unseen,lu2022mean}, we evaluate \methodname\ on two real-world tabletop datasets: the Object Clutter Indoor Dataset (OCID)~\cite{suchi2019easylabel} and the Object Segmentation Dataset (OSD)~\cite{richtsfeld2012segmentation}. \rui{To further assess generalization to photometrically challenging objects, we evaluate on HouseCat6D~\cite{jung2024housecat6d}, which includes reflective and transparent objects in complex backgrounds. We sample one frame every 10 frames across all HouseCat6D scenes, including its original training, validation, and test partitions, to evaluate transfer from TOD. All methods use the same evaluation image lists. A breakdown by HouseCat6D's official partitions in Appendix~\ref{app:housecat} confirms higher Overlap, Boundary, and Instance F-measures on the official test split across all three backbones.} Information of the \rui{three} datasets can be found in Tab.~\ref{tab:dataset}.

\begin{table}[t]
\caption{\rui{Statistics of the evaluation datasets.}}
\centering
\scalebox{0.82}{
    \begin{tabular}{|c|c|c|c|c|}
\hline
\textbf{Dataset} & \makecell{\textbf{Number of} \\ \rui{\textbf{Eval. Images}}} & \makecell{\textbf{Avg. Objects} \\ \textbf{per Image}} & \makecell{\textbf{Object Range} \\ \textbf{per Image}} & \makecell{\textbf{Number} \\ \textbf{of Objects}} \\ \hline
OCID~\cite{suchi2019easylabel}    & 2390 & 7.5 & 0--20 & 89 \\ \hline
OSD~\cite{richtsfeld2012segmentation} & 111 & 4.3 & 0--15 & - \\ \hline
HouseCat6D~\cite{jung2024housecat6d}  & 2503 & 6.7 & 5--10 & 194 \\ \hline
\end{tabular}

}
\label{tab:dataset}
\end{table}

\noindent\textbf{Metrics.} We evaluate UOIS performance using Precision (P), Recall (R), and F-measure (F) as outlined in \cite{xie2020best, xiang2020learning}. 
Specifically, we use the Hungarian algorithm to \rui{obtain a one-to-one assignment that maximizes the summed pairwise Overlap F-measure}.
The final precision, recall, and F-measure are computed as follows: $P=\frac{\sum_i\left|c_i \cap g\left(c_i\right)\right|}{\sum_i\left|c_i\right|}, R=\frac{\sum_i\left|c_i \cap g\left(c_i\right)\right|}{\sum_j\left|g_j\right|}, F=\frac{2PR}{P+R}$, where $c_i$ is the predicted object mask, $g\left(c_i\right)$ is the corresponding GT mask \rui{(empty for an unmatched prediction)}, and $g_j$ is the GT object. 
Overlap P/R/F assess segmentation accuracy based on the overlap between predicted and GT objects, while Boundary P/R/F evaluate \rui{boundary alignment with a tolerance of 0.003 times the image diagonal, rounded up to an integer number of pixels}.
Additionally, \rui{$\%75$ denotes the percentage of GT objects whose matched prediction has an Overlap F-measure strictly greater than $0.75$}~\cite{ochs2013segmentation}.
\rui{All UOIS segmentation metrics are computed per image and then averaged over the evaluation images; consequently, the reported F-measure is not necessarily the harmonic mean of the dataset-averaged precision and recall. Auxiliary prompt-level and candidate-level diagnostics follow their respective aggregation protocols as specified in each experiment and Appendix~\ref{app:hpg}.}
\rui{In the tables, bold and underlined values denote the best and second-best results in each column, respectively, where applicable.}

\noindent\textbf{Training.} \rui{We train HPG and \hdnetname\ on TOD~\cite{xie2020best}, using 4,000 scenes (10\% of its 40,000 training scenes) for the main experiments. We uniformly sample scenes and select the view with the most visible instances from each scene. The data percentages therefore refer to scene budgets.}
For a detailed analysis of the impact of different training data percentages, please refer to Sec.~\ref{sec:ablations}.

\rui{Our model is implemented in PyTorch.}
In \methodname, we use the ViT-L variant of SAM~\cite{kirillov2023segment} as the default promptable segmentation model and keep its pretrained parameters frozen. \rui{Throughout the experiments, we refer to this original Segment Anything Model~\cite{kirillov2023segment} as SAM1 (our default backbone for ablations and diagnostics), evaluate SAM2~\cite{ravi2024sam2} (SAM2.1 Hiera-L) as an upgraded frozen backbone, and compare against pretrained text-prompted SAM3~\cite{carion2026sam3} as an external baseline.} For the backbone of HPG, we employ the ViT-S variant of \rui{Depth Anything V2~\cite{depth_anything_v2}, keeping it frozen while training the HPG decoder, relative-depth fusion layers, and prediction heads}. The \rui{HPG} input images are resized to $H=336$ and $W=448$. 

\begin{table*}[t]
\caption{Segmentation results on OCID~\cite{suchi2019easylabel} and OSD~\cite{richtsfeld2012segmentation}. \rui{All scores are percentages.}}
\centering
\footnotesize
\setlength{\tabcolsep}{2.5pt}
\begin{adjustbox}{max width=\textwidth}

\begin{tabular}{|l|l|l|ccccccc|ccccccc|}
\hline
\multirow{3}{*}{Method} 
& \multirow{3}{*}{Input} 
& \multirow{3}{*}{\makecell{Training \\ Data}} 
& \multicolumn{7}{c|}{OCID} 
& \multicolumn{7}{c|}{OSD} \\ 
\cline{4-17} 
& & 
& \multicolumn{3}{c|}{Overlap} 
& \multicolumn{3}{c|}{Boundary} 
& \multirow{2}{*}{\%75}
& \multicolumn{3}{c|}{Overlap} 
& \multicolumn{3}{c|}{Boundary} 
& \multirow{2}{*}{\%75} \\ 
& & 
& \multicolumn{1}{c}{P} 
& \multicolumn{1}{c}{R} 
& \multicolumn{1}{c|}{F} 
& \multicolumn{1}{c}{P} 
& \multicolumn{1}{c}{R} 
& \multicolumn{1}{c|}{F} 
&
& \multicolumn{1}{c}{P} 
& \multicolumn{1}{c}{R} 
& \multicolumn{1}{c|}{F} 
& \multicolumn{1}{c}{P} 
& \multicolumn{1}{c}{R} 
& \multicolumn{1}{c|}{F} 
& \\ 
\hline

Mask R-CNN~\cite{he2017mask}
& RGB & TOD
& 77.6 & 67.0 & \multicolumn{1}{l|}{67.2}
& 65.5 & 53.9 & \multicolumn{1}{l|}{54.6}
& 55.8
& 64.2 & 61.3 & \multicolumn{1}{l|}{62.5}
& 50.2 & 40.2 & \multicolumn{1}{l|}{44.0}
& 31.9
\\

UCN~\cite{xiang2020learning}
& RGB & TOD
& 54.8 & 76.0 & \multicolumn{1}{l|}{59.4}
& 34.5 & 45.0 & \multicolumn{1}{l|}{36.5}
& 48.0
& 57.2 & 73.8 & \multicolumn{1}{l|}{63.3}
& 34.7 & 50.0 & \multicolumn{1}{l|}{39.1}
& 52.5
\\

Mask2Former~\cite{cheng2022masked}
& RGB & TOD
& 67.2 & 73.1 & \multicolumn{1}{l|}{67.1}
& 55.9 & 58.1 & \multicolumn{1}{l|}{54.5}
& 54.3
& 60.6 & 60.2 & \multicolumn{1}{l|}{59.5}
& 48.2 & 41.7 & \multicolumn{1}{l|}{43.3}
& 32.4
\\

UOAIS-Net~\cite{back2022unseen}
& RGB$\ast$ & UOAIS-SIM
& 66.5 & 83.1 & \multicolumn{1}{l|}{67.9}
& 62.1 & 70.2 & \multicolumn{1}{l|}{62.3}
& 73.1
& 84.2 & 83.7 & \multicolumn{1}{l|}{83.8}
& 72.2 & 72.8 & \multicolumn{1}{l|}{72.1}
& 76.7
\\

MSMFormer~\cite{lu2022mean}
& RGB & UOAIS-SIM
& 70.2 & 84.4 & \multicolumn{1}{l|}{70.5}
& 64.5 & 74.9 & \multicolumn{1}{l|}{64.9}
& 75.3
& 59.3 & 82.0 & \multicolumn{1}{l|}{67.9}
& 42.9 & 72.0 & \multicolumn{1}{l|}{52.4}
& 72.4
\\

MSMFormer~\cite{lu2022mean}
& RGB & TOD
& 72.9 & 68.3 & \multicolumn{1}{l|}{67.7}
& 60.5 & 56.3 & \multicolumn{1}{l|}{55.8}
& 52.9
& 63.4 & 64.7 & \multicolumn{1}{l|}{63.6}
& 48.6 & 47.4 & \multicolumn{1}{l|}{47.0}
& 40.2
\\

SAM3~\cite{carion2026sam3}
& RGB & $\circ$
& \textbf{96.4} & 82.9 & \multicolumn{1}{l|}{84.3}
& \textbf{94.1} & 79.8 & \multicolumn{1}{l|}{81.5}
& 85.6
& 83.7 & \underline{91.6} & \multicolumn{1}{l|}{86.8}
& 74.4 & \textbf{90.9} & \multicolumn{1}{l|}{80.3}
& \textbf{92.3}
\\

\hline

Mask R-CNN~\cite{he2017mask}
& Depth & TOD
& 85.3 & 85.6 & \multicolumn{1}{l|}{84.7}
& 83.2 & 76.6 & \multicolumn{1}{l|}{78.8}
& 72.7
& 77.8 & 85.1 & \multicolumn{1}{l|}{80.6}
& 52.5 & 57.9 & \multicolumn{1}{l|}{54.6}
& 77.6
\\

UOAIS-Net~\cite{back2022unseen}
& Depth$\ast$ & UOAIS-SIM
& 89.9 & 90.9 & \multicolumn{1}{l|}{89.8}
& 86.7 & \underline{84.1} & \multicolumn{1}{l|}{\underline{84.7}}
& 87.1
& 84.9 & 86.4 & \multicolumn{1}{l|}{85.5}
& 68.2 & 66.2 & \multicolumn{1}{l|}{66.9}
& 80.8
\\

FD-TNN~\cite{fu2024taylor}
& Depth & TOD
& 90.0 & 89.9 & \multicolumn{1}{l|}{89.7}
& 86.4 & 83.1 & \multicolumn{1}{l|}{84.4}
& -
& 88.0 & 87.4 & \multicolumn{1}{l|}{87.7}
& 74.3 & 72.8 & \multicolumn{1}{l|}{73.4}
& -
\\

\hline

UOAIS-Net~\cite{back2022unseen}
& RGB-D$\ast$ & UOAIS-SIM
& 70.7 & 86.7 & \multicolumn{1}{l|}{71.9}
& 68.2 & 78.5 & \multicolumn{1}{l|}{68.8}
& 78.7
& 85.3 & 85.4 & \multicolumn{1}{l|}{85.2}
& 72.7 & 74.3 & \multicolumn{1}{l|}{73.1}
& 79.1
\\

UCN~\cite{xiang2020learning}
& RGB-D & TOD
& 86.0 & \textbf{92.3} & \multicolumn{1}{l|}{88.5}
& 80.4 & 78.3 & \multicolumn{1}{l|}{78.8}
& 82.2
& 84.3 & 88.3 & \multicolumn{1}{l|}{86.2}
& 67.5 & 67.5 & \multicolumn{1}{l|}{67.1}
& 79.3
\\

Mask R-CNN~\cite{he2017mask}
& RGB-D & TOD
& 67.6 & 83.7 & \multicolumn{1}{l|}{68.9}
& 63.3 & 72.6 & \multicolumn{1}{l|}{64.1}
& 68.4
& 83.9 & 84.2 & \multicolumn{1}{l|}{79.1}
& 71.2 & 72.0 & \multicolumn{1}{l|}{70.9}
& 77.8
\\

Mask2Former~\cite{cheng2022masked}
& RGB-D & TOD
& 78.6 & 82.8 & \multicolumn{1}{l|}{79.5}
& 69.3 & 76.2 & \multicolumn{1}{l|}{71.1}
& 69.3
& 75.6 & 79.2 & \multicolumn{1}{l|}{77.3}
& 54.1 & 64.0 & \multicolumn{1}{l|}{58.0}
& 65.2
\\

UOIS-Net-2D~\cite{xie2021unseen}
& RGB-D & TOD
& 88.3 & 78.9 & \multicolumn{1}{l|}{81.7}
& 82.0 & 65.9 & \multicolumn{1}{l|}{71.4}
& 69.1
& 80.7 & 80.5 & \multicolumn{1}{l|}{79.9}
& 66.0 & 67.1 & \multicolumn{1}{l|}{65.6}
& 71.9
\\

UOIS-Net-3D~\cite{xie2021unseen}
& RGB-D & TOD
& 86.5 & 86.6 & \multicolumn{1}{l|}{86.4}
& 80.0 & 73.4 & \multicolumn{1}{l|}{76.2}
& 77.2
& 85.7 & 82.5 & \multicolumn{1}{l|}{83.3}
& 75.7 & 68.9 & \multicolumn{1}{l|}{71.2}
& 73.8
\\

MSMFormer~\cite{lu2022mean}
& RGB-D & UOAIS-SIM
& 78.2 & 83.8 & \multicolumn{1}{l|}{76.6}
& 69.1 & 78.4 & \multicolumn{1}{l|}{69.6}
& 75.3
& 80.6 & 86.9 & \multicolumn{1}{l|}{83.1}
& 65.9 & 72.9 & \multicolumn{1}{l|}{68.7}
& 80.2
\\

MSMFormer~\cite{lu2022mean}
& RGB-D & TOD
& 88.4 & 90.2 & \multicolumn{1}{l|}{88.5}
& 84.7 & 83.1 & \multicolumn{1}{l|}{83.0}
& 80.3
& 79.5 & 86.4 & \multicolumn{1}{l|}{82.8}
& 53.5 & 71.0 & \multicolumn{1}{l|}{60.6}
& 79.4
\\

ZISVFM~\cite{zisvfm}
& RGB-D & $\circ$
& 92.5 & 86.1 & \multicolumn{1}{l|}{89.2}
& 84.5 & 82.4 & \multicolumn{1}{l|}{83.5}
& 87.0
& 86.4 & 73.1 & \multicolumn{1}{l|}{78.1}
& 61.7 & 58.5 & \multicolumn{1}{l|}{60.0}
& 66.0
\\

\hline

UOIS-SAM (SAM1)
& RGB & TOD (10\%)
& \underline{92.8} & 88.6 & \multicolumn{1}{l|}{\underline{90.2}}
& 86.4 & 83.7 & \multicolumn{1}{l|}{84.4}
& \underline{88.7}
& \textbf{90.6} & 89.4 & \multicolumn{1}{l|}{\underline{90.0}}
& \underline{80.3} & 87.1 & \multicolumn{1}{l|}{\underline{83.1}}
& 88.9
\\
UOIS-SAM (SAM2)
& RGB & TOD (10\%)
& 91.6 & \underline{91.1} & \multicolumn{1}{l|}{\textbf{90.7}}
& \underline{89.0} & \textbf{86.7} & \multicolumn{1}{l|}{\textbf{87.0}}
& \textbf{89.8}
& \underline{90.4} & \textbf{92.0} & \multicolumn{1}{l|}{\textbf{91.1}}
& \textbf{84.0} & \underline{89.6} & \multicolumn{1}{l|}{\textbf{86.2}}
& \underline{90.9}
\\

\hline
\end{tabular}
\end{adjustbox}
\par\smallskip
\begin{minipage}{\textwidth}
\footnotesize
\rui{\textit{Note:} ``$\ast$'' denotes amodal supervision. ``$\circ$'' denotes no task-specific training. ``-'' denotes unavailable data.}
\end{minipage}
\label{tab:main_results}
\end{table*}

\begin{table}[t]
\caption{\rui{Segmentation results on HouseCat6D~\cite{jung2024housecat6d}. All scores are percentages.}}
\centering
\small
\setlength{\tabcolsep}{4pt}
\begin{adjustbox}{max width=\linewidth}

\begin{tabular}{|l|l|l|ccccccc|}
\hline
\multirow{3}{*}{Method} & \multirow{3}{*}{Input} & \multirow{3}{*}{\makecell{Training\\Data}} & \multicolumn{7}{c|}{HouseCat6D} \\
\cline{4-10}
& & & \multicolumn{3}{c|}{Overlap} & \multicolumn{3}{c|}{Boundary} & \multirow{2}{*}{\%75} \\
& & & P & R & \multicolumn{1}{c|}{F} & P & R & \multicolumn{1}{c|}{F} & \\
\hline
UCN~\cite{xiang2020learning}
& RGB
& TOD
& 34.7 & 76.2 & \multicolumn{1}{l|}{45.0}
& 18.0 & 35.7 & \multicolumn{1}{l|}{22.5}
& 48.4
\\
UOAIS-Net~\cite{back2022unseen}
& RGB$\ast$
& UOAIS-SIM
& 51.3 & 84.6 & \multicolumn{1}{l|}{60.3}
& 47.1 & 67.3 & \multicolumn{1}{l|}{52.8}
& 81.2
\\
MSMFormer~\cite{lu2022mean}
& RGB
& UOAIS-SIM
& 58.7 & 84.8 & \multicolumn{1}{l|}{67.3}
& 51.0 & 69.9 & \multicolumn{1}{l|}{57.6}
& 80.4
\\

SAM3~\cite{carion2026sam3}
& RGB & $\circ$
& 75.1 & 84.5 & \multicolumn{1}{l|}{77.4}
& 68.9 & \underline{82.6} & \multicolumn{1}{l|}{72.6}
& \textbf{94.9}
\\

\hline
UCN~\cite{xiang2020learning}
& RGB-D
& TOD
& 68.4 & 34.9 & \multicolumn{1}{l|}{35.5}
& 41.6 & 25.1 & \multicolumn{1}{l|}{18.5}
& 22.9
\\
UOAIS-Net~\cite{back2022unseen}
& RGB-D$\ast$
& UOAIS-SIM
& 52.0 & 84.1 & \multicolumn{1}{l|}{61.5}
& 40.1 & 60.1 & \multicolumn{1}{l|}{45.9}
& 76.5
\\
MSMFormer~\cite{lu2022mean}
& RGB-D
& UOAIS-SIM
& 16.7 & 60.9 & \multicolumn{1}{l|}{25.3}
& 6.4 & 24.3 & \multicolumn{1}{l|}{9.9}
& 17.1
\\
ZISVFM~\cite{zisvfm}
& RGB-D & $\circ$
& 58.5 & 19.4 & \multicolumn{1}{l|}{25.7}
& 25.9 & 9.2 & \multicolumn{1}{l|}{11.9}
& 12.5
\\
\hline
UOIS-SAM (SAM1)
& RGB
& TOD (10\%)
& \textbf{80.2}
& \underline{87.2}
& \multicolumn{1}{l|}{\textbf{81.8}}
& \textbf{77.4}
& 78.6
& \multicolumn{1}{l|}{\underline{76.5}}
& 86.8
\\
UOIS-SAM (SAM2)
& RGB & TOD (10\%)
& \underline{76.4} & \textbf{91.6} & \multicolumn{1}{l|}{\underline{81.4}}
& \underline{75.1} & \textbf{83.3} & \multicolumn{1}{l|}{\textbf{77.3}}
& \underline{91.8}
\\
\hline
\end{tabular}

\end{adjustbox}
\par\smallskip
\begin{minipage}{\linewidth}
\footnotesize
\rui{\textit{Note:} $^\ast$ denotes amodal supervision. ``$\circ$'' denotes no task-specific training.}
\end{minipage}
\label{tab:photometric_results}
\end{table}

\rui{We train HPG and \hdnetname\ \textbf{separately}.} Both networks are trained for 30 epochs, \rui{using batches of 32 images for HPG and a budget of $M=30$ point prompts per image for \hdnetname}. The real-world datasets are reserved for final zero-shot evaluation. The training process for both networks utilizes the AdamW optimizer~\cite{loshchilov2017decoupled}, with a step learning rate scheduler that decreases the learning rate by a factor of 0.1 every 10 epochs, starting from an initial learning rate of 1e-3. We apply a weight decay of 1e-4, with $\beta_1 = 0.9$ and $\beta_2 = 0.999$. No data augmentation is applied during the training of either network.

\noindent\textbf{Inference.}
\rui{RGB images are processed at $480\times640$ resolution. HPG generates a scene-adaptive number of point prompts using a foreground probability threshold of $0.90$ and a heatmap threshold of $0.012$. For SAM1, OLQA retains the highest-scoring mask among the three candidates generated per prompt, and the mask logits are binarized at $0.05$. Candidates are filtered using a minimum adapted quality score of $0.25$, a minimum overlap fraction of $0.75$ with the predicted foreground, and an area range of $[50,40000]$ pixels. In line with robotic manipulation pipelines that apply explicit foreground filtering to suppress background false positives~\cite{danielczuk2019segmenting,back2022unseen,zisvfm}, we adopt these conservative foreground thresholds by default.

For GCSI, we set $(\tau_u,\tau_s)=(0.04,0.55)$ and $(\lambda_u,\lambda_s,\lambda_c)=(0.75,1.10,0.35)$. A candidate is rejected if its IoU or IoS with any previously selected mask reaches $0.60$ or $0.85$, respectively. Otherwise, it is accepted when $\Delta_i\geq0.35$, with a maximum of 64 instances. Final instances are retained only if they contain at least 30 unassigned pixels and preserve at least 15\% of their original mask area. These selection settings are shared across the three evaluated datasets and across the SAM1, Semantic-SAM, and SEEM variants. The complete parameter specification, single-factor weight sensitivity, and quality--coverage operating regimes of GCSI are detailed in Appendix~\ref{app:gcsi}.}

\subsection{Comparison with State-of-the-Art Methods}
\label{sec:main_results}
\rui{We evaluate \methodname\ with two frozen promptable segmentation models, SAM ViT-L~\cite{kirillov2023segment} and SAM2.1 Hiera-L~\cite{ravi2024sam2}. For SAM2, we retrain \hdnetname\ on the TOD dataset using the same training protocol and shared HPG and GCSI, while keeping the segmentation backbone frozen. Neither variant uses target-domain training or adaptation.}

\noindent\textbf{Baselines.} 
\rui{Tabs.~\ref{tab:main_results} and~\ref{tab:photometric_results} compare \methodname\ with RGB, depth, and RGB-D baselines and report their corresponding training data. Mask R-CNN~\cite{he2017mask} and Mask2Former~\cite{cheng2022masked} are general-purpose segmentation models. UCN~\cite{xiang2020learning} and MSMFormer~\cite{lu2022mean} perform instance grouping using mean shift clustering and its transformer-based formulation, respectively. UOIS-Net-2D/3D~\cite{xie2021unseen} combine depth-based instance segmentation with RGB refinement, whereas UOAIS-Net~\cite{back2022unseen} models hierarchical amodal perception. FD-TNN~\cite{fu2024taylor} performs depth-only segmentation with a Taylor neural network, while ZISVFM~\cite{zisvfm} leverages SAM1 for zero-shot instance segmentation. For OCID and OSD, we report the results published in the corresponding papers. For HouseCat6D, we evaluate UCN, UOAIS-Net, and MSMFormer using the authors' released checkpoints and default settings without target-domain fine-tuning. For UOAIS-Net, we additionally apply a foreground segmentation network~\cite{wu2020cgnet} to filter background predictions. For SAM3~\cite{carion2026sam3}, which supports point prompting and text-conditioned segmentation, we evaluate its pretrained text-conditioned variant with the generic language prompt \texttt{item}, without task-specific weight updates.}

\noindent\textbf{Results on OCID and OSD.}
\rui{Tab.~\ref{tab:main_results} compares \methodname\ against representative RGB, depth, and RGB-D baselines. The SAM2 RGB-only variant achieves the highest Overlap and Boundary F-measures on both datasets. The SAM1 variant achieves Overlap/Boundary F-measures of 90.17\%/84.41\% on OCID and 89.95\%/83.12\% on OSD, while attaining the highest Overlap Precision on OSD (90.64\%). The SAM2 variant further improves the F-measures to 90.72\%/87.02\% on OCID and 91.09\%/86.24\% on OSD. Compared with the strong SAM3 baseline, SAM2 improves Overlap/Boundary F-measures by 6.4/5.5 pp on OCID and 4.3/5.9 pp on OSD. On OSD, SAM2 exceeds the best RGB-D results by 4.89 and 13.14 percentage points (pp) in Overlap and Boundary F-measures, respectively. These results demonstrate that \methodname\ achieves strong segmentation performance using RGB input and limited synthetic supervision, without target-domain fine-tuning.}

\noindent\textbf{Results on HouseCat6D.}
\rui{As shown in Tab.~\ref{tab:photometric_results}, both \methodname\ variants achieve higher Overlap and Boundary F-measures than the compared RGB and RGB-D baselines without target-domain fine-tuning. The SAM1 variant achieves 81.77\% Overlap F and 76.51\% Boundary F, compared with 77.4\% and 72.6\% for SAM3, respectively. It also attains the highest Overlap Precision (80.16\%) and Boundary Precision (77.44\%). The SAM2 variant achieves 81.38\% Overlap F and 77.27\% Boundary F, with 91.77\% on $\%75$; SAM3 attains the highest $\%75$ score (94.9\%). These results show strong overall RGB generalization in cluttered scenes with reflective and transparent objects. Category diagnostics in Appendix~\ref{app:hpg} show that the shared conservative foreground filter lowers final recall on transparent glassware and thin cutlery to suppress background false positives despite high pre-filtering candidate availability.}

\subsection{Ablation Studies}
\label{sec:ablations}
\rui{All ablations use SAM1 ViT-L and TOD supervision. We report Overlap/Boundary F-measure, and $\%75$ for final segmentation, and use separate diagnostics to examine output instances.}

\begin{table}[t]
\caption{\rui{Effect of the main components on final segmentation. All scores are percentages.}}
\centering
\footnotesize
\setlength{\tabcolsep}{2pt}
\begin{adjustbox}{max width=\linewidth}
\begin{tabular}{|c|c|c|cc|c|cc|c|cc|c|}
\hline
\multirow{3}{*}{HPG}
& \multirow{3}{*}{\hdnetname}
& \multirow{3}{*}{GCSI}
& \multicolumn{3}{c|}{OCID}
& \multicolumn{3}{c|}{OSD}
& \multicolumn{3}{c|}{HouseCat6D} \\
\cline{4-12}

& &
& \multicolumn{2}{c|}{F-measure}
& \multirow{2}{*}{\%75}
& \multicolumn{2}{c|}{F-measure}
& \multirow{2}{*}{\%75}
& \multicolumn{2}{c|}{F-measure}
& \multirow{2}{*}{\%75} \\

& &
& Overlap & Boundary &
& Overlap & Boundary &
& Overlap & Boundary & \\

\hline

& & 
& 39.58 & 33.02 & 78.14
& 37.40 & 29.25 & 74.10
& 35.82 & 27.36 & \underline{88.51}
\\

& & \cmark
& 42.66 & 40.76 & 79.52
& 39.57 & 34.20 & 72.13
& 38.60 & 31.97 & 86.26
\\

\cmark & &
& 80.09 & 67.93 & 82.71
& 76.80 & 60.48 & 79.12
& 72.65 & 63.20 & 86.89
\\

\cmark & & \cmark
& 82.38 & 75.39 & 79.46
& 78.17 & 68.84 & 68.77
& 75.07 & 68.78 & 78.90
\\

\cmark & \cmark &
& \underline{88.15} & \underline{82.09} & \textbf{90.35}
& \underline{88.61} & \underline{82.24} & \textbf{89.97}
& \underline{78.47} & \underline{74.11} & \textbf{95.07}
\\

\cmark & \cmark & \cmark
& \textbf{90.17} & \textbf{84.41} & \underline{88.73}
& \textbf{89.95} & \textbf{83.12} & \underline{88.94}
& \textbf{81.77} & \textbf{76.51} & 86.80
\\

\hline
\end{tabular}
\end{adjustbox}
\label{tab:components_results}
\end{table}

\noindent\textbf{Effect of Main Components.}
\rui{Tab.~\ref{tab:components_results} evaluates the contributions of HPG, \hdnetname, and GCSI using frozen SAM1. Without HPG, a fixed $32\times32$ grid of point prompts is used; without \hdnetname, the original SAM quality scores are retained; and NMS is applied when GCSI is disabled. HPG substantially improves Overlap F-measure from 39.58\%, 37.40\%, and 35.82\% to 80.09\%, 76.80\%, and 72.65\% on OCID, OSD, and HouseCat6D, respectively. Adding \hdnetname\ further increases these scores to 88.15\%, 88.61\%, and 78.47\%. GCSI improves them to 90.17\%, 89.95\%, and 81.77\%, with corresponding gains in Boundary F-measure. 
GCSI also reduces the fraction of GT objects recovered above the \%75 criterion, reflecting the object coverage cost of producing a more coherent instance set.}

\noindent\textbf{\rui{Effect of HPG.}}
\rui{We compare HPG with three prompting strategies using the same number of prompts per image. Uniform distributes points across the entire image, matching the number generated by HPG for each image. FG-Random randomly samples points from the predicted foreground, while FG-Spatial distributes points within the same foreground without heatmap guidance. Prompt purity measures the percentage of prompts on GT foreground, while coverage measures the percentage of GT instances receiving at least one prompt.

As shown in Tab.~\ref{tab:hpg_ablation}, sampling within the predicted foreground substantially improves prompt purity, object coverage, and segmentation accuracy over Uniform. HPG achieves the highest purity, Overlap F-measure, and Boundary F-measure on all three datasets. Compared with FG-Spatial, HPG improves purity by 1.13--2.38 pp and Overlap F-measure by 0.84--1.52 pp, while maintaining comparable object coverage. These results demonstrate that foreground prediction improves prompt allocation, while heatmap guidance provides further benefits beyond foreground sampling alone. On HouseCat6D, using DAV2 features with relative depth further improves prompt coverage and pre-filtering candidate recall on transparent and specular objects while preserving high recall on the remaining categories. Please refer to Appendix~\ref{app:hpg} for this feature ablation alongside HPG threshold sensitivity, category-level retention bottlenecks, prompt-budget saturation, and latency scaling.}

\begin{table}[t]
\caption{Effect of different prompt sampling strategies.}
\centering
\footnotesize
\setlength{\tabcolsep}{3pt}
\begin{adjustbox}{max width=\columnwidth}
\begin{tabular}{|l|l|cc|cc|}
\hline
\multirow{2}{*}{Dataset}
& \multirow{2}{*}{Method}
& \multicolumn{2}{c|}{Prompt Allocation}
& \multicolumn{2}{c|}{F-measure} \\
\cline{3-6}
& & Purity $\uparrow$ & Coverage $\uparrow$
& Overlap $\uparrow$ & Boundary $\uparrow$ \\
\hline

\multirow{4}{*}{OCID}
& Uniform
& 19.68 & 54.36 & 63.79 & 56.16 \\
& FG-Random
& \underline{95.65} & 88.91 & 88.06 & 81.68 \\
& FG-Spatial
& 95.34 & \textbf{95.82}
& \underline{89.33} & \underline{83.43} \\
& HPG
& \textbf{97.72} & \underline{95.20}
& \textbf{90.17} & \textbf{84.41} \\
\hline

\multirow{4}{*}{OSD}
& Uniform
& 17.28 & 81.86 & 77.49 & 70.38 \\
& FG-Random
& \underline{93.03} & \underline{98.52}
& \underline{89.10} & \underline{81.82} \\
& FG-Spatial
& 92.78 & \textbf{99.37}
& 88.43 & 81.57 \\
& HPG
& \textbf{93.91} & \underline{98.52}
& \textbf{89.95} & \textbf{83.12} \\
\hline

\multirow{4}{*}{HouseCat6D}
& Uniform
& 12.76 & 67.29 & 69.64 & 63.70 \\
& FG-Random
& 67.52 & 93.08
& 80.56 & 75.76 \\
& FG-Spatial
& \underline{67.62} & \underline{96.45}
& \underline{80.80} & \underline{76.12} \\
& HPG
& \textbf{68.96} & \textbf{96.59}
& \textbf{81.77} & \textbf{76.51} \\
\hline
\end{tabular}
\end{adjustbox}
\label{tab:hpg_ablation}
\end{table}

\begin{table*}[t]
\caption{\rui{Comparison of adaptation methods. Segmentation scores are percentages.}}
\centering
\footnotesize
\setlength{\tabcolsep}{5pt}
\begin{adjustbox}{max width=\textwidth}
\begin{tabular}{|l|cc|c|cc|c|ccc|}
\hline
\multirow{3}{*}{\makecell[l]{Adaptation\\Methods}}
& \multicolumn{3}{c|}{OCID}
& \multicolumn{3}{c|}{OSD}
& \multicolumn{3}{c|}{Training Efficiency} \\
\cline{2-10}

& \multicolumn{2}{c|}{F-measure}
& \multirow{2}{*}{\%75}
& \multicolumn{2}{c|}{F-measure}
& \multirow{2}{*}{\%75}
& \multirow{2}{*}{\makecell{Trainable\\Params (M)}}
& \multirow{2}{*}{\makecell{Throughput\\(images/s)}}
& \multirow{2}{*}{\makecell{Peak Mem.\\(GiB)}} \\

& Overlap & Boundary &
& Overlap & Boundary &
& & & \\

\hline

Original IoU head
& 82.38 & 75.39 & 79.46
& 78.17 & 68.84 & 68.77
& 0 & -- & -- \\

Mask decoder tuning
& 88.04 & 82.23 & 88.89
& 85.95 & 79.56 & 82.82
& 4.058 & 5.077 & 4.485 \\

LoRA~\cite{hulora} + Mask decoder tuning
& 87.70 & 82.61 & 85.14
& 86.32 & 80.89 & 80.90
& 4.845 & 1.974 & 5.769 \\

SAM-Adapter~\cite{sam_adapter}
& 87.07 & 81.59 & 83.70
& 86.08 & 80.02 & 80.38
& 4.175 & 2.037 & 5.795 \\

IoU head tuning
& 87.68 & 81.16 & \textbf{89.63}
& 87.19 & 80.22 & 88.14
& 0.133 & 5.695 & 4.305 \\

\hline

\hdnetname\
& \textbf{90.17} & \textbf{84.41} & 88.73
& \textbf{89.95} & \textbf{83.12} & \textbf{88.94}
& 0.197 & 5.679 & 4.306 \\

\hline
\end{tabular}
\end{adjustbox}
\par\smallskip
\label{tab:candidate_adaptation}
\end{table*}

\noindent\textbf{Comparison of Different Adaptation Methods.}
\rui{Tab.~\ref{tab:candidate_adaptation} compares adaptation strategies under the same HPG and GCSI setting and training protocol. The Original IoU head uses pretrained SAM quality scores. Mask decoder tuning optimizes the decoder, while LoRA~\cite{hulora} and SAM-Adapter~\cite{sam_adapter} adapt the image encoder together with decoder tuning. IoU head tuning updates only the original SAM IoU prediction head.

\hdnetname\ achieves the highest Overlap and Boundary F-measures on both datasets, reaching 90.17\%/84.41\% on OCID and 89.95\%/83.12\% on OSD. Compared with IoU head tuning, these metrics improve by 2.49/3.25 and 2.76/2.90 pp, respectively. \hdnetname\ uses only 0.197M trainable parameters, compared with at least 4.058M for decoder adaptation, while maintaining training throughput and memory usage comparable to IoU head tuning. These results demonstrate the effectiveness and efficiency of adapting candidate quality scores without updating the pretrained mask generator.}

\begin{table}[t]
\caption{\rui{Mechanism decomposition of GCSI on SAM1. All scores are percentages.}}
\centering
\footnotesize
\setlength{\tabcolsep}{2.5pt}
\begin{adjustbox}{max width=\columnwidth}
\begin{tabular}{|l|l|ccccc|}
\hline
Dataset
& Method
& \makecell{Overlap\\F $\uparrow$}
& \makecell{Boundary\\F $\uparrow$}
& \makecell{Instance\\F $\uparrow$}
& \makecell{FP\\Rate $\downarrow$}
& \makecell{Miss\\Rate $\downarrow$} \\
\hline

\multirow{4}{*}{OCID}
& NMS & 88.15 & 82.09 & 88.77 & 11.84 & \textbf{7.88} \\
& Prefiltered NMS & 90.16 & 84.27 & 91.43 & 6.19 & 9.53 \\
& Hard gates only & 90.16 & 84.23 & 91.40 & 6.30 & 9.51 \\
& Full GCSI & \textbf{90.17} & \textbf{84.41} & \textbf{91.75} & \textbf{5.54} & 9.57 \\
\hline

\multirow{4}{*}{OSD}
& NMS & 88.61 & 82.24 & 82.30 & 22.37 & \textbf{8.51} \\
& Prefiltered NMS & 89.94 & 83.06 & 85.37 & 17.67 & 9.40 \\
& Hard gates only & \textbf{90.01} & \textbf{83.28} & 85.48 & 17.50 & 9.40 \\
& Full GCSI & 89.95 & 83.12 & \textbf{85.53} & \textbf{17.40} & 9.40 \\
\hline

\multirow{4}{*}{HouseCat6D}
& NMS & 78.47 & 74.11 & 80.00 & 28.87 & \textbf{4.27} \\
& Prefiltered NMS & 81.75 & 76.37 & 81.65 & 20.69 & 12.59 \\
& Hard gates only & 81.73 & 76.32 & 81.67 & 20.67 & 12.58 \\
& Full GCSI & \textbf{81.77} & \textbf{76.51} & \textbf{82.08} & \textbf{19.97} & 12.60 \\
\hline
\end{tabular}
\end{adjustbox}
\label{tab:cross_prompt_selection}
\end{table}

\noindent\textbf{\rui{Effect of GCSI.}}
\rui{Tab.~\ref{tab:cross_prompt_selection} breaks down GCSI on SAM1 over a shared frozen candidate pool on final non-overlapping label maps (Instance F uses one-to-one matching at IoU $\geq 0.5$). Over NMS, Full GCSI raises Instance F by 2.09--3.23\,pp and cuts FP Rate by 4.97--8.90\,pp across all three datasets while raising Miss Rate (4.27\% to 12.60\% on HouseCat6D). Adding GCSI's admission filters ($\tau_{\mathrm{fg,ov}}$, score, area) and composition to NMS (\emph{Prefiltered NMS}) drives most of the Overlap F gain (+1.33--3.28\,pp). Compared with \emph{Hard gates only} ($\lambda_u=\lambda_s=\lambda_c=0$), adding soft penalties (\emph{Full GCSI}) lowers FP by 0.10--0.75\,pp and raises Instance F by 0.05--0.41\,pp across all datasets (+0.18/+0.19\,pp Boundary F on OCID/HouseCat6D). Thus, filtering and composition drive Overlap F, whereas GCSI's selector improves instance coherence and FP control; Appendix~\ref{app:gcsi} reports the full three-architecture decomposition (with larger soft-penalty gains on SEEM) and $\tau_{\mathrm{fg,ov}}$ coverage trade-offs.}

\subsection{Analysis}
\begin{table}[t]
\caption{Candidate scoring with fixed prompts and SAM masks.}
\centering
\footnotesize
\setlength{\tabcolsep}{2pt}
\begin{adjustbox}{max width=\columnwidth}
\begin{tabular}{|l|ccc|ccc|}
\hline
\multirow{2}{*}{Dataset}
& \multicolumn{3}{c|}{Mean Selected IoU $\uparrow$}
& \multicolumn{3}{c|}{Object Recall@0.5 (\%) $\uparrow$} \\
\cline{2-7}
& \makecell{Original\\IoU Head}
& \hdnetname\
& Oracle
& \makecell{Original\\IoU Head}
& \hdnetname\
& Oracle \\
\hline

OCID
& 0.6454 & 0.8555 & 0.8890
& 90.15 & 91.92 & 94.69 \\

OSD
& 0.6714 & 0.8733 & 0.9067
& 92.41 & 93.67 & 95.78 \\

HouseCat6D
& 0.7346 & 0.8905 & 0.9146
& 94.04 & 95.97 & 96.48 \\

\hline
\end{tabular}
\end{adjustbox}
\label{tab:candidate_scoring}
\end{table}

\noindent\textbf{\rui{Candidate Scoring Diagnostics.}}
\rui{Tab.~\ref{tab:candidate_scoring} isolates candidate scoring by keeping HPG prompts and SAM candidate masks fixed. The Original IoU head uses pretrained SAM quality scores, while \hdnetname\ uses adapted scores. Oracle selects the candidate with the highest GT IoU for each foreground prompt. Mean Selected IoU measures mask quality over foreground prompts, while Object Recall@0.5 measures the percentage of GT objects with at least one selected candidate reaching IoU $\geq 0.5$ before candidate filtering and GCSI.}

\rui{\hdnetname\ substantially improves candidate selection on all three datasets, recovering 85.8--86.6\% of the gap between the Original IoU head and Oracle in Mean Selected IoU. Object Recall@0.5 also improves without changing the candidate masks. Oracle achieves Object Recall@0.5 of 94.69\%, 95.78\%, and 96.48\%, indicating that HPG and frozen SAM already provide viable candidates for most GT objects. 
Together, these results highlight the complementary roles of HPG in object discovery and \hdnetname\ in selecting accurate masks from the available candidates.}

\begin{table}[t]
\caption{Effect of TOD training data size on OCID.}
\centering
\footnotesize
\setlength{\tabcolsep}{5pt}
\begin{adjustbox}{max width=\columnwidth}
\begin{tabular}{|c|r|cc|c|}
\hline
\multirow{2}{*}{\makecell{TOD\\Scenes}}
& \multirow{2}{*}{\makecell{No. of\\Scenes}}
& \multicolumn{2}{c|}{F-measure}
& \multirow{2}{*}{\%75} \\
\cline{3-4}
& & Overlap & Boundary & \\
\hline

1\%
& 400
& 86.59 & 80.50 & 89.38 \\

5\%
& 2,000
& 88.60 & 82.94 & \textbf{91.35} \\

10\%
& 4,000
& \textbf{90.17} & 84.41 & 88.73 \\

50\%
& 20,000
& 90.06 & \textbf{84.61} & 89.62 \\

100\%
& 40,000
& 89.89 & 84.38 & 88.73 \\
\hline
\end{tabular}
\end{adjustbox}
\label{tab:data}
\end{table}

\noindent\textbf{\rui{Effect of Training Data Size.}}
\rui{Tab.~\ref{tab:data} evaluates the effect of TOD training data size on OCID. With only 400 scenes (1\%), \methodname\ achieves 86.59\% Overlap F-measure and 80.50\% Boundary F-measure. Using 4,000 scenes (10\%) increases these scores to 90.17\% and 84.41\%, respectively. This setting achieves the highest Overlap F-measure among all training sizes, while its Boundary F-measure is only 0.20 pp below the best result obtained with 20,000 scenes. These results demonstrate strong data efficiency, with limited additional benefit from increasing the training data beyond 10\%.}

\begin{table}[t]
\caption{Evaluation across different promptable segmentation architectures.}
\centering
\footnotesize
\setlength{\tabcolsep}{6pt}
\renewcommand{\arraystretch}{1.08}
\begin{adjustbox}{max width=\linewidth}
\begin{tabular}{|c|c|c|cc|cc|cc|}
\hline
\multirow{3}{*}{HPG}
& \multirow{3}{*}{\hdnetname}
& \multirow{3}{*}{GCSI}
& \multicolumn{2}{c|}{OCID}
& \multicolumn{2}{c|}{OSD}
& \multicolumn{2}{c|}{HouseCat6D} \\
\cline{4-9}
& & &
\multicolumn{2}{c|}{F-measure}
& \multicolumn{2}{c|}{F-measure}
& \multicolumn{2}{c|}{F-measure} \\
\cline{4-9}
& & &
Overlap & Boundary
& Overlap & Boundary
& Overlap & Boundary \\
\hline

\multicolumn{9}{|l|}{\textbf{SAM1 ViT-L}~\cite{kirillov2023segment}} \\
\hline
& & 
& 39.58 & 33.02
& 37.40 & 29.25
& 35.82 & 27.36 \\
\cmark & &
& 80.09 & 67.93
& 76.80 & 60.48
& 72.65 & 63.20 \\
\cmark & \cmark &
& \underline{88.15} & \underline{82.09}
& \underline{88.61} & \underline{82.24}
& \underline{78.47} & \underline{74.11} \\
\cmark & \cmark & \cmark
& \textbf{90.17} & \textbf{84.41}
& \textbf{89.95} & \textbf{83.12}
& \textbf{81.77} & \textbf{76.51} \\
\hline

\multicolumn{9}{|l|}{\textbf{Semantic-SAM}~\cite{li2024segment}} \\
\hline
& &
& 60.53 & 61.82
& 42.41 & 37.52
& 38.70 & 34.84 \\
\cmark & &
& 85.96 & 80.81
& 79.96 & 71.92
& 78.86 & 73.25 \\
\cmark & \cmark &
& \underline{88.13} & \underline{81.99}
& \underline{87.04} & \underline{80.22}
& \underline{79.74} & \underline{74.13} \\
\cmark & \cmark & \cmark
& \textbf{88.39} & \textbf{82.36}
& \textbf{87.60} & \textbf{80.96}
& \textbf{80.56} & \textbf{74.39} \\
\hline

\multicolumn{9}{|l|}{\textbf{SEEM Focal-T}~\cite{zou2023segment}} \\
\hline
& &
& 50.20 & 39.06
& 31.01 & 21.33
& 35.58 & 30.18 \\
\cmark & &
& 78.65 & 63.46
& \underline{67.80} & 48.07
& 74.16 & 60.56 \\
\cmark & \cmark &
& \textbf{80.17} & \underline{65.05}
& \textbf{70.19} & \underline{50.19}
& \underline{75.22} & \underline{61.32} \\
\cmark & \cmark & \cmark
& \underline{79.98} & \textbf{65.12}
& \textbf{70.19} & \textbf{51.07}
& \textbf{75.76} & \textbf{61.47} \\
\hline
\end{tabular}
\end{adjustbox}
\label{tab:cross_arch_generalization}
\end{table}

\subsection{Generalization Across Promptable Segmentation Architectures}
\label{sec:cross_arch_generalization}
\rui{We evaluate whether \methodname\ generalizes across frozen SAM1 ViT-L, Semantic-SAM, and SEEM Focal-T, whose native scores reflect predicted mask IoU, query quality, and prompt--mask matching, respectively. Each architecture uses an independently trained \hdnetname\ on 10\% TOD scenes without shared adapter weights or target-domain tuning (details in Appendix~\ref{app:olqa}).}

\rui{Following the component protocol of Tab.~\ref{tab:components_results}, Tab.~\ref{tab:cross_arch_generalization} evaluates all three stages across nine model--dataset pairs. Both HPG and \hdnetname\ improve Overlap and Boundary F-measures in all 9/9 pairs, confirming that visible-object quality adaptation transfers across diverse native scoring mechanisms. Applying the same shared GCSI parameters improves Boundary F-measure in all 9/9 pairs and Overlap F-measure in 7/9 pairs (tied on SEEM--OSD and slightly lower on SEEM--OCID, 80.17\% to 79.98\%), demonstrating that the framework generalizes beyond SAM1.}

\begin{table*}[t]
\caption{Deployment costs.}
\label{tab:deployment_cost}
\centering
\footnotesize
\setlength{\tabcolsep}{4pt}

\begin{tabular}{@{}c@{\hspace{14pt}}c@{\hspace{14pt}}c@{}}

\begin{tabular}[t]{@{}lr@{}}
\multicolumn{2}{c}{\textbf{(a) Inference time}} \\[2pt]
\toprule
Method & Time (ms) \\
\midrule
UCN~\cite{xiang2020learning}             & 344.71 \\
MSMFormer~\cite{lu2022mean}               & 327.55 \\
UOAIS-Net~\cite{back2022unseen} & 170.51 \\
ZISVFM~\cite{zisvfm}                     & 6226.99 \\
\midrule
\methodname                               & 258.14 \\
\bottomrule
\end{tabular}

&

\begin{tabular}[t]{@{}lr@{}}
\multicolumn{2}{c}{\textbf{(b) Deployment resources}} \\[2pt]
\toprule
Resource & Value \\
\midrule
Total parameters         & 343.62M \\
Tuning parameters        & 6.49M \\
Model weights            & 1.39 GiB \\
Inference peak memory    & 4.47 GiB \\
Training peak memory     & 14.19 GiB \\
Training cost  & 2.13 GPU-h \\
\bottomrule
\end{tabular}

&

\begin{tabular}[t]{@{}lr@{}}
\multicolumn{2}{c}{\textbf{(c) Latency breakdown}} \\[2pt]
\toprule
Component & Time (ms) \\
\midrule
HPG                      & 42.77 \\
SAM image encoder        & 151.85 \\
Mask decoder and \hdnetname\   & 47.84 \\
GCSI                     & 15.67 \\
\midrule
Total                     & 258.14 \\
\bottomrule
\end{tabular}

\end{tabular}
\end{table*}

\subsection{Deployment Cost}
\label{sec:deployment_cost}
\rui{Tab.~\ref{tab:deployment_cost} compares the deployment cost of the complete \methodname\ with SAM1 ViT-L against representative UOIS methods. All inference times are measured on a single RTX 4090D GPU using FP32 and batch size 1. We process $480\times640$ images and produce instance maps at the same resolution, while each method retains its native internal processing resolution. Timing covers the complete pipeline from decoded CPU input to the final instance map.}

\rui{The profiled implementation achieves an average inference time of 258.14 ms (3.87 FPS), outperforming UCN and MSMFormer with refinement, which require 344.71 and 327.55 ms, respectively. The complete model contains 343.62M parameters, of which only 6.49M are used for task adaptation. Its model weights occupy 1.39 GiB, with a peak inference memory of 4.47 GiB.}

\rui{The SAM image encoder dominates inference time at 151.85 ms (58.8\%), followed by the mask decoder with \hdnetname\ (47.84 ms) and HPG (42.77 ms, including preprocessing and prompt generation). With vectorized mask statistics and reuse of the composed instance map, GCSI requires only 15.67 ms on average, accounting for 6.1\% of the total latency. The estimated training cost for HPG, \hdnetname, and SAM feature caching is 2.13 GPU-hours, excluding foundation model pretraining (whereas Tab.~\ref{tab:candidate_adaptation} profiles throughput with online SAM forward passes for fair comparison with decoder fine-tuning). Appendix~\ref{app:hpg} further profiles how each pipeline stage scales across naturally occurring scene prompt-count bins.}

\subsection{Suction Grasping Analysis on a Public Benchmark}
\label{sec:suction_analysis}

\begin{table}[t]
\caption{Segmentation and suction grasping results on 2,340 RealSense test images from SuctionNet-1Billion.}
\centering
\footnotesize
\setlength{\tabcolsep}{2.2pt}
\begin{adjustbox}{max width=\columnwidth}
\begin{tabular}{@{}lcc|cccc@{}}
\toprule
\multirow{2}{*}{Instance Masks}
& \multicolumn{2}{c|}{Segmentation}
& \multicolumn{4}{c}{Suction Grasping} \\
\cmidrule(lr){2-3}\cmidrule(l){4-7}
& \makecell{Overlap\\F $\uparrow$}
& \makecell{Boundary\\F $\uparrow$}
& \makecell{Scene\\AP@50 $\uparrow$}
& \makecell{Target\\AP@10 $\uparrow$}
& \makecell{Success\\@10 $\uparrow$}
& \makecell{Missing\\Targets $\downarrow$} \\
\midrule

UCN~\cite{xiang2020learning}
& 45.02 & 30.71
& 16.69 & 8.55 & 33.06 & 8,293 \\

MSMFormer~\cite{lu2022mean}
& 53.26 & 30.59
& 21.67 & 12.79 & 49.04 & 2,212 \\

UOAIS-Net~\cite{back2022unseen}
& 81.04 & 58.77
& 25.92 & 16.36 & 56.18 & 1,843 \\

\methodname\ (NMS)
& 85.10 & 69.24
& 26.25 & 17.78 & 61.53 & \textbf{585} \\

\methodname\ (GCSI)
& \textbf{88.84} & \textbf{72.30}
& \textbf{27.31} & \textbf{18.27} & \textbf{61.64} & 853 \\

\midrule
GT masks
& 100.00 & 100.00
& 28.34 & 19.04 & 63.80 & 0 \\
\bottomrule
\end{tabular}

\end{adjustbox}
\label{tab:downstream_suction}
\end{table}

\noindent\textbf{\rui{Evaluation Protocol.}}
\rui{We evaluate downstream suction grasping on 2,340 RealSense images sampled every 10 frames from 90 SuctionNet-1Billion~\cite{suctionnet} test scenes (19,261 visible target observations) using a shared frozen UISN~\cite{cao2024uncertainty} and varying only the input masks. UCN and MSMFormer use TOD RGB-D models without refinement, and UOAIS-Net uses RGB-D with CG-Net. Alongside standard Overlap/Boundary F-measures~\cite{xie2021unseen} and scene AP@50, we match predicted masks to GT targets via Hungarian IoU assignment and penalize candidates contacting a wrong object as incorrect. Across all 19,261 targets (scoring unmatched targets as zero), Target AP@10 averages prefix precision over ranks 1--10 and $q\in\{0.2,0.4,0.6,0.8\}$, Success@10 measures targets with a valid top-10 contact at $q\geq0.2$, and Missing Targets counts unmatched GT objects.}

\noindent\textbf{\rui{Suction Grasping Performance.}}
\rui{As shown in Tab.~\ref{tab:downstream_suction}, stronger instance segmentation generally corresponds to better suction performance across the evaluated methods. Among methods using predicted masks, \methodname\ with GCSI achieves the highest Overlap F, Boundary F, scene AP@50, Target AP@10, and Success@10. Compared with UOAIS-Net, the Overlap and Boundary F increase from 81.04\% and 58.77\% to 88.84\% and 72.30\%, respectively. Scene AP@50 increases from 25.92\% to 27.31\%, while Target AP@10 increases from 16.36\% to 18.27\%. Success@10 rises from 56.18\% to 61.64\%, corresponding to 1,051 additional successful target observations, while Missing Targets decreases from 1,843 to 853. 
Using GT masks yields 19.04\% Target AP@10 with the same frozen UISN. These results show that improved instance segmentation benefits both suction quality and the availability of valid suction candidates across objects.}

\noindent\textbf{\rui{Role of GCSI.}}
\rui{Compared with the NMS baseline, \methodname\ with GCSI improves scene AP@50 from 26.25\% to 27.31\% and Target AP@10 from 17.78\% to 18.27\%, while reducing the number of predicted instance masks by 13.9\%. The improvement in Target AP@10 mainly comes from more accurate grasp predictions for targets covered by both output instance sets. These results suggest that organizing instance masks into a compact and coherent set can benefit downstream grasp prediction beyond simply retaining more masks.}

\begin{figure}[t]
\centering
\includegraphics[width=\columnwidth]{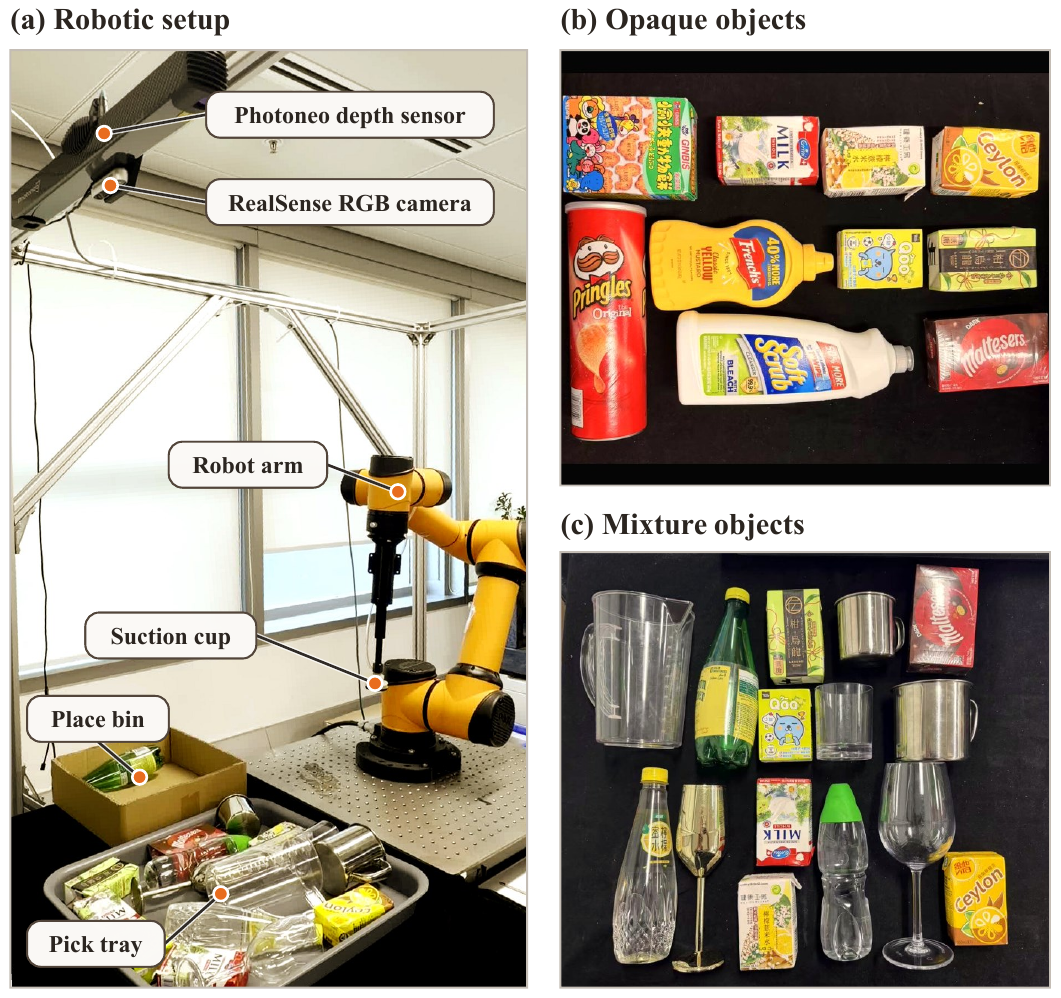}
\caption{Robot setup and evaluation objects. (a) Robot and cameras. (b) Opaque household objects. (c) A mixture of Opaque, transparent, and reflective objects.}
\label{fig:real_robot_setting}
\end{figure}

\subsection{Real-World Suction Grasping Evaluation}
\label{sec:real_suction}

We evaluate suction grasping in cluttered tabletop scenes using a robot equipped with a suction gripper (Fig.~\ref{fig:real_robot_setting}). An Intel RealSense camera provides RGB images, while a calibrated Photoneo scanner provides 3D point clouds. For transparent and reflective objects, UA-Depth~\cite{cao2026uadepth} restores missing or unreliable depth. The instance masks and corresponding 3D geometry are then used for suction prediction with UISN~\cite{cao2024uncertainty}.

We consider two settings: \emph{Opaque}, containing opaque household objects, and \emph{Mixture}, containing opaque, transparent, and reflective objects. \methodname\ performs instance segmentation from RGB images using GCSI. UOAIS-Net uses RGB-D with CG-Net in Opaque and a retrained RGB model without CG-Net \rui{(to avoid severe raw-depth corruption on transparent and reflective objects)} in Mixture. Both methods use the same UISN for suction prediction.

\begin{figure*}[th]
\centering
\includegraphics[width=\textwidth]{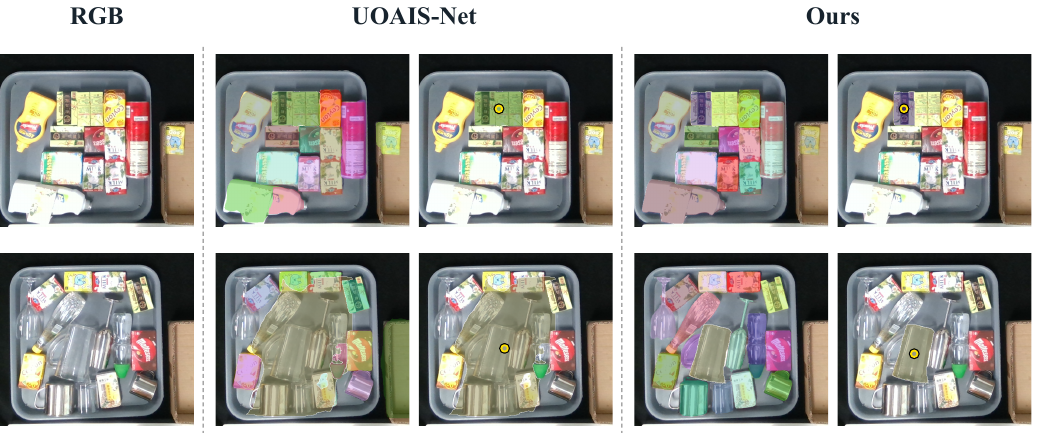}
\caption{Qualitative suction results in Opaque (top) and Mixture (bottom). Each method's predicted instances and selected suction point (yellow ring) are shown. \methodname\ separates neighboring objects that are merged by UOAIS-Net, including transparent objects in Mixture.}
\label{fig:real_qualitative}
\end{figure*}

\noindent\textbf{Evaluation Protocol.}
For each trial, we randomly select a point prompt generated by \methodname\ to specify the target object \rui{in the workspace without manual bias}. 
Both methods use the same point to select their corresponding predicted instance masks, ensuring that the same \rui{physical} target is evaluated. 
We record 225 paired target evaluations: 99 in Opaque and 126 in Mixture.
A robot trial is successful only when the suction gripper attaches to the target object, lifts it, and deposits it into the bin; all other executions are failures. 
In the rare event of a motion planning failure, we repeat the trial. 

\begin{table}[t]
\caption{Suction success rates on paired targets.}
\label{tab:real_suction}
\centering
\footnotesize
\setlength{\tabcolsep}{4pt}
\begin{adjustbox}{max width=\columnwidth}
\begin{tabular}{llcc}
\toprule
Setting & Method & Successes / Total & Success Rate (\%) \\
\midrule
Opaque & UOAIS-Net & 83/99 & 83.8 \\
 & \methodname\ & \textbf{92/99} & \textbf{92.9} \\
\midrule
Mixture & UOAIS-Net & 63/126 & 50.0 \\
 & \methodname\ & \textbf{115/126} & \textbf{91.3} \\
\bottomrule
\end{tabular}
\end{adjustbox}
\end{table}

\noindent\textbf{Results.}
As shown in Tab.~\ref{tab:real_suction}, \methodname\ achieves success rates of 92.9\% in Opaque and 91.3\% in Mixture, compared with 83.8\% and 50.0\% for UOAIS-Net, respectively. The corresponding improvements are 9.1 and 41.3 percentage points. Fig.~\ref{fig:real_qualitative} illustrates how separating neighboring objects helps preserve the intended target during suction-point selection. \rui{Representative suction grasping executions are provided in the supplementary video.}

\begin{table}[t]
\caption{Failure categories in the paired evaluation. Each entry reports Opaque / Mixture / Total counts.}
\label{tab:real_failure_modes}
\centering
\footnotesize
\setlength{\tabcolsep}{4pt}
\begin{adjustbox}{max width=\columnwidth}
\begin{tabular}{lcc}
\toprule
Failure category & UOAIS-Net & \methodname \\
\midrule
Adjacent objects merged & 5 / 5 / 10 & 1 / 0 / 1 \\
Background included in mask & 5 / 46 / 51 & 0 / 0 / 0 \\
Partial object mask & 0 / 0 / 0 & 1 / 0 / 1 \\
Incorrect suction point & 4 / 4 / 8 & 4 / 3 / 7 \\
Depth or contact geometry & 1 / 8 / 9 & 0 / 7 / 7 \\
Transfer or placement & 1 / 0 / 1 & 1 / 1 / 2 \\
\midrule
Total & 16 / 63 / 79 & 7 / 11 / 18 \\
\bottomrule
\end{tabular}
\end{adjustbox}
\end{table}

\noindent\textbf{Failure Analysis.}
Tab.~\ref{tab:real_failure_modes} groups each unsuccessful trial by its dominant failure. Segmentation and target association errors account for 61 of 79 UOAIS-Net failures, compared with 2 of 18 for \methodname. Most UOAIS-Net segmentation failures arise from masks that include background regions, including the supporting tray. For \methodname, 14 of the 18 failures concern suction point selection or depth and contact geometry. These results show that accurate instance segmentation substantially reduces target identification errors, leaving suction point selection and contact estimation as the main remaining challenges.

\section{Conclusion}
\rui{We presented \methodname, a data-efficient framework that adapts frozen promptable segmentation models for robotic UOIS.} HPG provides foreground-aware point prompts, \hdnetname\ adapts candidate quality estimates, and GCSI constructs the final instance set. Experiments on OCID, OSD, and HouseCat6D demonstrate strong RGB-based segmentation with limited synthetic supervision. Evaluations with SAM1, Semantic-SAM, and SEEM further illustrate the applicability of the framework to different promptable segmentation architectures, \rui{while downstream evaluations on SuctionNet-1Billion and 225 paired real-world robotic suction trials---including cluttered scenes with transparent and reflective objects---confirm substantial gains in robotic manipulation.}

\noindent\textbf{\rui{Limitations and Future Work.}}
\rui{First, \methodname\ predicts visible (modal) rather than amodal masks, which may limit downstream planning in heavily stacked clutter. Second, as diagnosed in Appendix~\ref{app:hpg}, the shared conservative foreground filter suppresses background false positives across benchmarks, but it can reduce recall on transparent glassware and thin cutlery even when accurate candidates are available prior to filtering. Although relaxing the foreground thresholds recovers recall at inference without retraining, developing adaptive and confidence-calibrated foreground filtering remains an important direction. Third, HPG and \hdnetname\ are currently trained separately with fixed inference thresholds. Exploring joint end-to-end training and amodal extensions is left for future work.}

\clearpage
\section*{Appendix Overview}

The following appendices provide additional implementation details, hyperparameter sensitivity analyses, and diagnostic evaluations:
(i)~Appendix~\ref{app:gcsi} details the complete GCSI parameter specification, a controlled mechanism decomposition across all three architectures, independent soft-penalty weight sensitivity, and quality--coverage operating trade-offs;
(ii)~Appendix~\ref{app:hpg} examines HPG inference threshold sensitivity, backbone feature and relative-depth ablations, category-level pipeline retention bottlenecks, prompt-budget scaling, and measured runtime scaling;
(iii)~Appendix~\ref{app:housecat} reports HouseCat6D results broken down by official dataset partitions;
(iv)~Appendix~\ref{app:olqa} documents the exact cross-architecture feature taps and design details of \hdnetname; and
(v)~Appendix~\ref{app:realworld} summarizes the hardware setup, sensing pipeline, and paired trial filtering for the real-world suction grasping evaluation.
The default benchmark and cross-architecture evaluations use the shared production configuration reported in the main paper without target-domain tuning, whereas the mechanism decompositions and sensitivity analyses vary the targeted parameters as noted in each section. Unless otherwise noted, segmentation metrics are computed per image and averaged over each dataset (OCID: 2,390 images; OSD: 111 images; HouseCat6D: 2,503 sampled images), while prompt purity, prompt coverage, and Final Instance Recall@0.5 (one-to-one matched after GCSI, distinct from the pre-filtering candidate Object Recall@0.5 in the main paper) are reported as micro ratios over total prompts or ground-truth (GT) instances.

\appendices
\setcounter{table}{0}
\renewcommand{\thetable}{A-\Roman{table}}
\setcounter{figure}{0}
\renewcommand{\thefigure}{A-\arabic{figure}}
\renewcommand{\theHtable}{appendix.\arabic{table}}
\renewcommand{\theHfigure}{appendix.\arabic{figure}}

\section{GCSI Parameter Specification, Sensitivity, and Quality--Coverage Trade-off}
\label{app:gcsi}

\subsection{Complete Production Configuration}
To aid reproducibility and clarify how each hyperparameter acts within the consolidation pipeline, Table~\ref{tab:app_gcsi_params} groups all shared production GCSI parameters into four functional stages: candidate admission, hard conflict gating, soft redundancy scoring, and non-overlapping visibility composition. These settings are fixed across all three datasets and across SAM1, Semantic-SAM, and SEEM.

\begin{table}[!t]
\caption{Shared production GCSI parameters grouped by functional stage.}
\label{tab:app_gcsi_params}
\centering
\footnotesize
\setlength{\tabcolsep}{4pt}
\renewcommand{\arraystretch}{1.04}
\begin{adjustbox}{max width=\linewidth}
\begin{tabular}{lllr}
\toprule
Stage & Symbol & Parameter Description & Value \\
\midrule
\multirow{5}{*}{Admission}
 & $\tau_{\mathrm{mask}}$ & Mask logit binarization threshold & 0.05 \\
 & $s_{\min}$ & Minimum adapted quality score & 0.25 \\
 & $\tau_{\mathrm{fg,ov}}$ & Min.\ candidate overlap with pred.\ FG & 0.75 \\
 & $a_{\min}$ & Minimum candidate mask area (px) & 50 \\
 & $a_{\max}$ & Maximum candidate mask area (px) & 40,000 \\
\midrule
\multirow{3}{*}{Hard Gate}
 & $\tau_u^{\mathrm{hard}}$ & Hard pairwise IoU rejection threshold & 0.60 \\
 & $\tau_s^{\mathrm{hard}}$ & Hard pairwise IoS rejection threshold & 0.85 \\
 & $\tau_{\mathrm{dup}}$ & Exact duplicate guard (IoU / IoS) & 0.999 \\
\midrule
\multirow{7}{*}{\makecell[l]{Soft Penalty\\\& Utility}}
 & $\tau_u$ & Soft IoU activation threshold & 0.04 \\
 & $\tau_s$ & Soft IoS activation threshold & 0.55 \\
 & $\lambda_u$ & Soft IoU penalty weight & 0.75 \\
 & $\lambda_s$ & Soft IoS penalty weight & 1.10 \\
 & $\lambda_c$ & Containment penalty weight & 0.35 \\
 & $\tau_{\Delta}$ & Minimum marginal utility threshold & 0.35 \\
 & $N_{\max}$ & Maximum selected instances per image & 64 \\
\midrule
\multirow{2}{*}{Composition}
 & $a_{\mathrm{vis}}$ & Min.\ unassigned visible area (px) & 30 \\
 & $f_{\mathrm{vis}}$ & Min.\ preserved visible fraction & 0.15 \\
\bottomrule
\end{tabular}
\end{adjustbox}
\end{table}

\begin{table*}[!t]
\caption{Controlled GCSI mechanism decomposition across three promptable architectures (\%).}
\label{tab:app_gcsi_decomposition}
\centering
\scriptsize
\setlength{\tabcolsep}{2.2pt}
\renewcommand{\arraystretch}{1.04}
\begin{adjustbox}{max width=\textwidth}
\begin{tabular}{llrrrrrrrrrrrrrrr}
\toprule
 & & \multicolumn{5}{c}{OCID ($N=2{,}390$)} & \multicolumn{5}{c}{OSD ($N=111$)} & \multicolumn{5}{c}{HouseCat6D ($N=2{,}503$)} \\
\cmidrule(lr){3-7}\cmidrule(lr){8-12}\cmidrule(lr){13-17}
Architecture & Method & OF $\uparrow$ & BF $\uparrow$ & IF $\uparrow$ & FP $\downarrow$ & Miss $\downarrow$ & OF $\uparrow$ & BF $\uparrow$ & IF $\uparrow$ & FP $\downarrow$ & Miss $\downarrow$ & OF $\uparrow$ & BF $\uparrow$ & IF $\uparrow$ & FP $\downarrow$ & Miss $\downarrow$ \\
\midrule
\multirow{4}{*}{SAM1 ViT-L}
 & NMS & 88.15 & 82.09 & 88.77 & 11.84 & \textbf{7.88} & 88.61 & 82.24 & 82.30 & 22.37 & \textbf{8.51} & 78.47 & 74.11 & 80.00 & 28.87 & \textbf{4.27} \\
 & Prefiltered NMS & 90.16 & 84.27 & 91.43 & 6.19 & 9.53 & 89.94 & 83.06 & 85.37 & 17.67 & 9.40 & 81.75 & 76.37 & 81.65 & 20.69 & 12.59 \\
 & Hard gates only & 90.16 & 84.23 & 91.40 & 6.30 & 9.51 & \textbf{90.01} & \textbf{83.28} & 85.48 & 17.50 & 9.40 & 81.73 & 76.32 & 81.67 & 20.67 & 12.58 \\
 & Full GCSI & \textbf{90.17} & \textbf{84.41} & \textbf{91.75} & \textbf{5.54} & 9.57 & 89.95 & 83.12 & \textbf{85.53} & \textbf{17.40} & 9.40 & \textbf{81.77} & \textbf{76.51} & \textbf{82.08} & \textbf{19.97} & 12.60 \\
\midrule
\multirow{4}{*}{Semantic-SAM}
 & NMS & 88.13 & 81.99 & 89.13 & 9.06 & \textbf{10.68} & 87.04 & 80.22 & 81.44 & 20.56 & \textbf{12.88} & 79.74 & 74.13 & 80.59 & 24.57 & \textbf{9.83} \\
 & Prefiltered NMS & \textbf{88.71} & \textbf{82.53} & \textbf{89.56} & 7.63 & 11.55 & 87.57 & \textbf{80.96} & \textbf{82.66} & \textbf{18.72} & 12.94 & \textbf{80.57} & 74.27 & 80.49 & 21.21 & 14.47 \\
 & Hard gates only & 88.46 & 82.39 & 89.28 & 7.87 & 11.99 & 87.59 & 80.93 & 82.58 & 18.81 & 13.00 & 80.52 & 74.20 & 80.40 & 21.29 & 14.58 \\
 & Full GCSI & 88.39 & 82.36 & 89.45 & \textbf{7.35} & 12.13 & \textbf{87.60} & \textbf{80.96} & 82.61 & 18.76 & 13.00 & 80.56 & \textbf{74.39} & \textbf{80.76} & \textbf{20.63} & 14.63 \\
\midrule
\multirow{4}{*}{SEEM Focal-T}
 & NMS & 80.17 & 65.05 & 76.45 & 22.60 & \textbf{21.39} & 70.19 & 50.19 & 50.56 & 55.02 & \textbf{37.03} & 75.22 & 61.32 & 74.53 & 32.19 & \textbf{14.00} \\
 & Prefiltered NMS & \textbf{80.18} & 64.83 & 75.68 & 23.71 & 21.93 & \textbf{70.44} & 50.07 & 50.46 & 55.11 & 37.09 & 75.70 & 60.93 & 73.85 & 30.49 & 18.28 \\
 & Hard gates only & 79.90 & 64.47 & 75.25 & 24.23 & 22.14 & 69.62 & 49.32 & 49.64 & 55.81 & 37.75 & 75.50 & 60.50 & 73.17 & 31.41 & 18.59 \\
 & Full GCSI & 79.98 & \textbf{65.12} & \textbf{76.78} & \textbf{21.08} & 22.47 & 70.19 & \textbf{51.07} & \textbf{51.95} & \textbf{52.59} & 38.25 & \textbf{75.76} & \textbf{61.47} & \textbf{74.92} & \textbf{28.34} & 18.77 \\
\bottomrule
\end{tabular}
\end{adjustbox}
\end{table*}

\subsection{Controlled Mechanism Decomposition Across Architectures}
To isolate how candidate admission, hard conflict gates, and soft redundancy penalties each contribute across promptable models, Table~\ref{tab:app_gcsi_decomposition} evaluates all variants on the same frozen top-1 candidate pools across all three backbones and benchmarks. All methods are evaluated on final non-overlapping 2D instance label maps (Instance F uses one-to-one matching at $\mathrm{IoU}\ge 0.5$, while FP Rate and Miss Rate denote the fractions of unmatched predicted instances and unmatched GT instances, respectively), and except for \emph{NMS}, all variants share identical default candidate admission filtering and mask composition. Uncertainty is assessed using 2,000 paired bootstrap resamples clustered by OCID sequence, HouseCat6D scene, and OSD image.

\noindent\textbf{Admission prefiltering vs.\ selector contribution.}
Comparing \emph{NMS} to \emph{Prefiltered NMS} isolates the combined effect of default candidate admission ($\tau_{\mathrm{fg,ov}}=0.75$, $s_{\min}=0.25$, area bounds) and greedy non-overlapping mask composition: this drives the primary Overlap F gains (+2.02, +1.33, and +3.28\,pp on SAM1) while reducing FP by 1.43--8.18\,pp on SAM1 and Semantic-SAM. Under matched prefiltering and composition, \emph{Full GCSI} vs.\ \emph{Prefiltered NMS} yields mixed Overlap F changes across architectures (positive in 5/9 pairs, including +0.01--0.02\,pp on SAM1, while $-0.32$\,pp [$-0.58, -0.11$] on Semantic-SAM--OCID and $-0.21$/$-0.25$\,pp on SEEM--OCID/OSD), but improves Boundary F in 8/9 pairs at unrounded precision (with Semantic-SAM--OSD at $+0.001$\,pp, tied at two-decimal table precision, and Semantic-SAM--OCID at $-0.17$\,pp), Instance F in 7/9 pairs, and lowers FP in 8/9 pairs. Gains are largest on SEEM's noisier candidate pools: +0.29/+1.00/+0.54\,pp Boundary F, +1.10/+1.48/+1.08\,pp Instance F, and $-2.63$/$-2.52$/$-2.15$\,pp FP on OCID/OSD/HouseCat6D. Note that \emph{Prefiltered NMS} vs.\ \emph{Hard gates only} compares two distinct selectors (standard IoU NMS at $0.20$/$0.30$ vs.\ hard IoU/IoS gates at $0.60$/$0.85$ with marginal utility cutoff $\tau_\Delta=0.35$).

\noindent\textbf{Joint necessity of soft redundancy penalties.}
While one-at-a-time sweeps appear flat when the remaining soft weights stay active (Sec.~\ref{app:gcsi_weights}), disabling all three soft penalties simultaneously (\emph{Hard gates only}, $\lambda_u=\lambda_s=\lambda_c=0$, $\tau_\Delta=0.35$) degrades performance relative to \emph{Full GCSI}: enabling soft penalties lowers FP and increases Instance F in \textbf{all 9/9} architecture--dataset pairs (SAM1: $-0.75$/$-0.10$/$-0.70$\,pp FP, $+0.35$/$+0.05$/$+0.41$\,pp IF; Semantic-SAM: $-0.52$/$-0.04$/$-0.66$\,pp FP, $+0.18$/$+0.02$/$+0.35$\,pp IF; SEEM: $-3.15$/$-3.23$/$-3.07$\,pp FP, $+1.52$/$+2.30$/$+1.75$\,pp IF) and improves Boundary F in 7/9 pairs (up to +1.74\,pp [1.25, 2.23] on SEEM--OSD), with small Overlap/Boundary F dips only on SAM1--OSD ($-0.06$/$-0.16$\,pp) and Semantic-SAM--OCID ($-0.07$/$-0.02$\,pp). Indeed, without soft penalties, the looser hard gates underperform \emph{Prefiltered NMS} on SEEM (e.g., OSD BF 49.32\% vs.\ 50.07\%), confirming that hard gates and soft penalties operate as a coupled suppression stage. Finally, disabling the final visibility thresholds ($a_{\mathrm{vis}}=0$, $f_{\mathrm{vis}}=0$) while retaining non-overlapping mask composition changes at most 0--9 images per dataset across all models, because upstream hard and soft suppression already reduces candidates dropped during mask composition from 16--2,310 (\emph{Prefiltered NMS}) to 0--9 (\emph{Full GCSI}).

\begin{figure*}[!t]
\centering
\includegraphics[width=\textwidth]{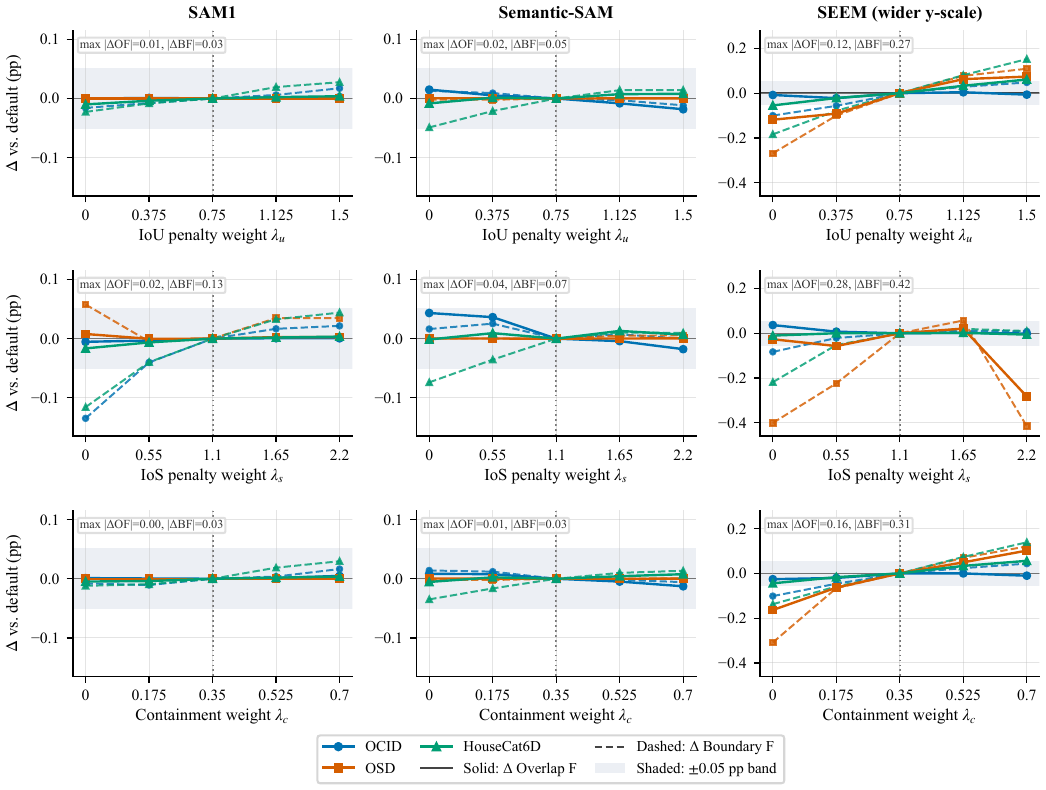}
\caption{Independent sensitivity of the three GCSI soft redundancy weights ($\lambda_u$, $\lambda_s$, $\lambda_c$) varied from $0\times$ to $2\times$ their default values across SAM1, Semantic-SAM, and SEEM on OCID, OSD, and HouseCat6D (117 configurations total). Solid curves show $\Delta$ Overlap F and dashed curves show $\Delta$ Boundary F relative to the default production setting (vertical dotted line), with hard conflict gates held active. Shaded regions denote a $\pm 0.05$\,pp reference band; note the wider vertical scale for SEEM.}
\label{fig:app_gcsi_weights}
\end{figure*}

\subsection{Independent Sensitivity of Soft Redundancy Weights}
\label{app:gcsi_weights}
To assess robustness to the soft penalty weights without conflating them with candidate filtering, we vary each weight ($\lambda_u$, $\lambda_s$, $\lambda_c$) independently across $\{0\times, 0.5\times, 1\times, 1.5\times, 2\times\}$ its default value while holding all candidate pools, admission filters, hard conflict gates, and composition rules fixed. Fig.~\ref{fig:app_gcsi_weights} summarizes all 117 model--dataset--weight combinations.

\noindent\textbf{Wide stability plateau on strong backbones.}
Across the entire $[0\times, 2\times]$ sweep, SAM1 exhibits mean variations of at most $|\Delta\text{OF}| \le 0.02$\,pp and $|\Delta\text{BF}| \le 0.13$\,pp; Semantic-SAM similarly stays within $|\Delta\text{OF}| \le 0.04$\,pp and $|\Delta\text{BF}| \le 0.07$\,pp. Mechanistically, this stability arises because the hard conflict gates already intercept most gross duplicates before soft penalties are evaluated (rejecting an average of 35.2, 38.4, and 43.3 overlapping proposals per image on OCID, OSD, and HouseCat6D for SAM1), leaving the soft redundancy terms to resolve only partially overlapping candidates.

\noindent\textbf{Soft terms remain behaviorally active.}
Small aggregate means do not imply that soft penalties are inactive. Disabling the soft IoS term ($\lambda_s=0$) alters the final instance label map on 151 OCID images and 164 HouseCat6D images for SAM1, increasing false positives by 0.59\,pp and 0.53\,pp and reducing Boundary F by 0.13\,pp and 0.12\,pp, respectively. Thus, soft penalties act as a targeted refinement mechanism for partially overlapping masks without destabilizing overall performance.

\noindent\textbf{Sensitivity across backbones.}
SEEM produces noisier candidate pools with substantially more borderline overlaps (e.g., 4.35 active soft IoU and 2.26 active soft IoS pairs per image on OSD), resulting in higher weight sensitivity (right column of Fig.~\ref{fig:app_gcsi_weights}). In particular, doubling $\lambda_s$ from 1.10 to 2.20 on SEEM--OSD reduces Overlap F by 0.28\,pp and Boundary F by 0.42\,pp while increasing the miss rate by 0.45\,pp. Similarly, disabling containment ($\lambda_c=0$) on SEEM--OSD lowers Overlap F by 0.16\,pp even though miss rate slightly decreases ($-0.19$\,pp). Keeping $\lambda_s$ near the moderate default avoids excessive suppression on noisier architectures.

\begin{figure*}[!t]
\centering
\includegraphics[width=0.92\textwidth]{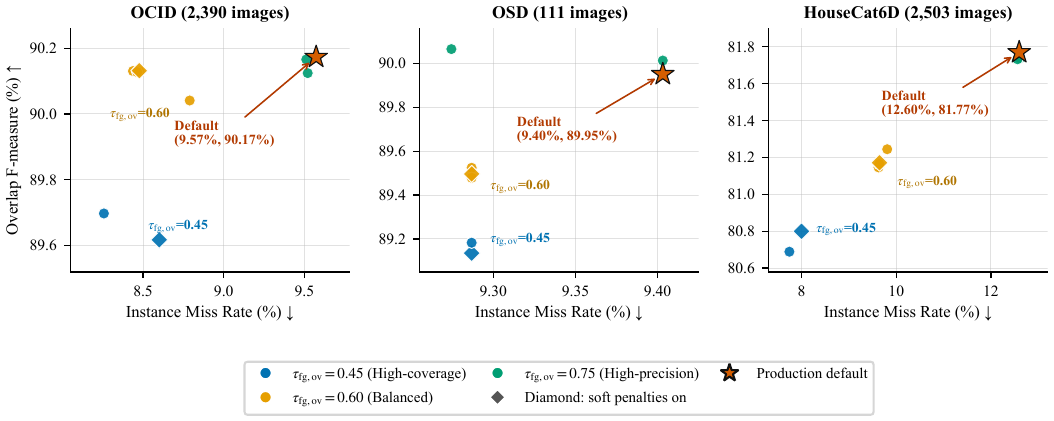}
\caption{Quality--coverage trade-off (Overlap F vs.\ instance miss rate) for SAM1 across 10 joint GCSI operating configurations on OCID, OSD, and HouseCat6D. Colors denote the candidate foreground overlap threshold $\tau_{\mathrm{fg,ov}} \in \{0.45, 0.60, 0.75\}$, diamonds indicate active soft redundancy penalties, and the star marks the shared production default.}
\label{fig:app_gcsi_pareto}
\end{figure*}

\subsection{Quality--Coverage Operating Regimes}
To clarify the coverage trade-off noted when enforcing global instance coherence, Fig.~\ref{fig:app_gcsi_pareto} plots 10 joint GCSI operating points spanning three foreground overlap regimes ($\tau_{\mathrm{fg,ov}} \in \{0.45, 0.60, 0.75\}$), quality cutoffs ($s_{\min} \in \{0.25, 0.40, 0.55\}$), and soft penalty settings.

\noindent\textbf{Controllable precision--coverage knob.}
Across all three datasets, the operating points separate cleanly by $\tau_{\mathrm{fg,ov}}$:
\begin{itemize}
    \item \emph{High-coverage regime ($\tau_{\mathrm{fg,ov}}=0.45$):} On HouseCat6D, SAM1 miss rate drops from 12.60\% to 7.74--8.00\% (raising $\%75$ from 86.80\% to 91.33--91.55\%), at the cost of lower Overlap F (80.69--80.80\%) and higher false positives (23.03--24.41\% vs.\ 19.97\%).
    \item \emph{Balanced regime ($\tau_{\mathrm{fg,ov}}=0.60$):} Yields intermediate coverage (9.63--9.81\% miss on HouseCat6D; 8.44--8.79\% on OCID) with 81.15--81.25\% (HouseCat6D) and 90.04--90.13\% (OCID) Overlap F.
    \item \emph{High-precision regime ($\tau_{\mathrm{fg,ov}}=0.75$):} Favors high segmentation accuracy and suppression of background false positives across datasets, and includes our shared production default, making it well suited for manipulation tasks where spurious background masks harm grasp selection.
\end{itemize}
Semantic-SAM and SEEM exhibit the same three operating regimes: moving from $\tau_{\mathrm{fg,ov}}=0.45$ to $0.75$ improves HouseCat6D Overlap F from 79.49--79.72\% to 80.53--80.64\% (Semantic-SAM) and 74.99--75.63\% to 75.53--75.81\% (SEEM) while trading 4--5\,pp of instance recall for fewer false positives. Practitioners deploying UOIS-SAM in high-recall settings can therefore lower $\tau_{\mathrm{fg,ov}}$ to 0.45--0.60 without retraining.

\section{HPG Threshold Sensitivity, Bottleneck Diagnostics, and Budget/Runtime Scaling}
\label{app:hpg}

\begin{figure*}[!t]
\centering
\includegraphics[width=0.92\textwidth]{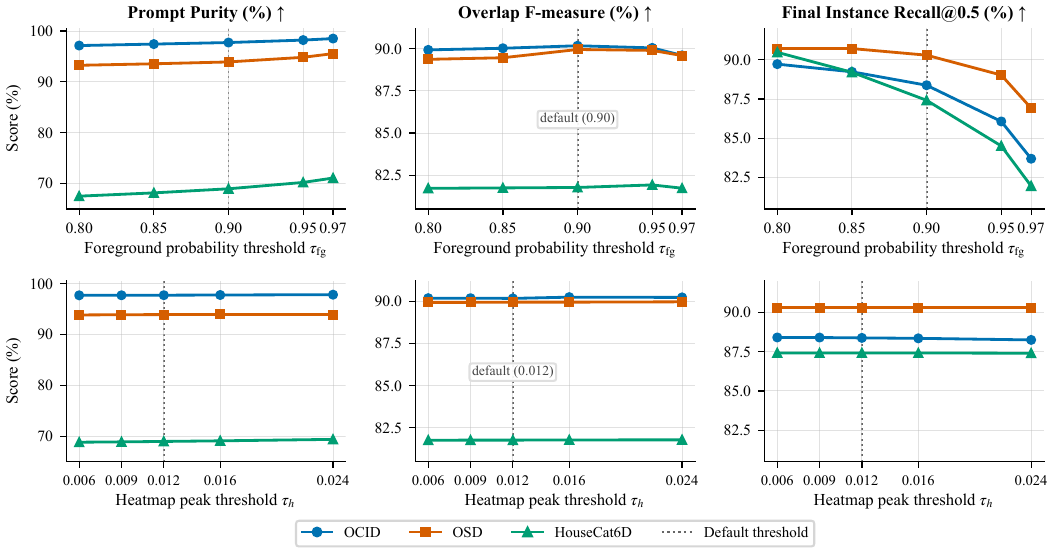}
\caption{Sensitivity of production UOIS-SAM (SAM1) to HPG inference thresholds on OCID, OSD, and HouseCat6D. Top row: foreground probability threshold $\tau_{\mathrm{fg}} \in [0.80, 0.97]$. Bottom row: heatmap peak threshold $\tau_h \in [0.006, 0.024]$. Matching vertical scales across rows highlight that performance is virtually invariant to $\tau_h$, whereas $\tau_{\mathrm{fg}}$ governs the purity--recall trade-off with Overlap F plateauing near the default $\tau_{\mathrm{fg}}=0.90$.}
\label{fig:app_hpg_thresholds}
\end{figure*}

\subsection{Inference Threshold Sensitivity}
HPG uses two inference thresholds: the pixel-level foreground probability threshold $\tau_{\mathrm{fg}}=0.90$ (which gates both prompt placement and the downstream candidate foreground mask) and the heatmap peak threshold $\tau_h=0.012$. Note that $\tau_{\mathrm{fg}}$ is distinct from both the candidate overlap requirement $\tau_{\mathrm{fg,ov}}=0.75$ and the $\%75$ evaluation metric. Fig.~\ref{fig:app_hpg_thresholds} sweeps each threshold independently across all three datasets under the full production pipeline.

\noindent\textbf{Invariance to heatmap threshold $\tau_h$.}
Varying $\tau_h$ across a $4\times$ interval $[0.006, 0.024]$ produces almost completely flat curves on all three datasets: Overlap F changes by at most $0.07$\,pp, Boundary F by at most $0.08$\,pp, and Final Instance Recall@0.5 by at most $0.13$\,pp. Hence, $\tau_h$ requires no domain-specific tuning.

\noindent\textbf{Purity--recall control via $\tau_{\mathrm{fg}}$.}
By contrast, $\tau_{\mathrm{fg}}$ smoothly trades prompt purity against instance recall. On HouseCat6D, raising $\tau_{\mathrm{fg}}$ from 0.90 to 0.97 increases prompt purity from 68.96\% to 71.07\% (FP: 19.97\% to 16.27\%) but lowers Final Instance Recall@0.5 from 87.42\% to 81.96\%. Lowering $\tau_{\mathrm{fg}}$ to 0.80 raises Final Instance Recall@0.5 to 90.48\% on HouseCat6D (+3.06\,pp), 89.72\% on OCID (+1.34\,pp), and 90.72\% on OSD (+0.42\,pp, two objects), but significantly increases FP on every benchmark: +2.56\,pp on HouseCat6D (22.53\%, 95\% CI [+2.08, +3.09]), +2.35\,pp on OSD (17.40\% to 19.75\%, CI [+0.98, +3.96]), and +1.02\,pp on OCID (5.54\% to 6.56\%, CI [+0.62, +1.45]). It also reduces Instance F across all three datasets ($-0.22$\,pp HouseCat6D, 82.08\% to 81.86\%; $-1.39$\,pp OSD, 85.53\% to 84.14\%; $-0.13$\,pp OCID, 91.75\% to 91.61\%), degrades OSD Overlap/Boundary F ($-0.58$/$-0.99$\,pp to 89.37\%/82.13\%), and worsens HouseCat6D count MAE from 1.68 to 2.05 (7.80 to 8.41 mean instances vs.\ 6.73 GT). Thus, $\tau_{\mathrm{fg}}=0.90$ sits on the Overlap F and Instance F plateau across benchmarks.

\begin{figure*}[!t]
\centering
\includegraphics[width=0.93\textwidth]{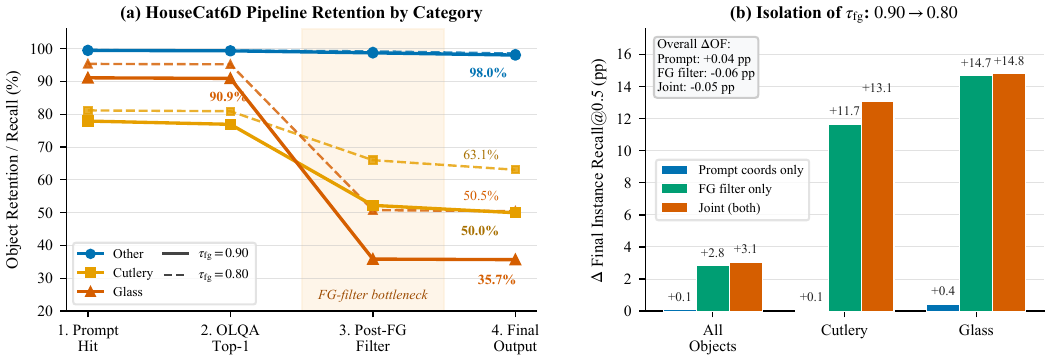}
\caption{Diagnostic analysis of object retention on HouseCat6D (SAM1 production pipeline). (a)~Micro object retention across four pipeline stages grouped by category proxy: Other (13,611 GT instances), Cutlery (1,603), and Glass (1,629)---covering 16,843 categorized GT instances out of 16,844 total---comparing default $\tau_{\mathrm{fg}}=0.90$ (solid) and relaxed $\tau_{\mathrm{fg}}=0.80$ (dashed). (b)~Counterfactual isolation of lowering $\tau_{\mathrm{fg}}$ from 0.90 to 0.80 on Final Instance Recall@0.5: changing only prompt locations vs.\ changing only the downstream foreground filter vs.\ changing both jointly.}
\label{fig:app_hpg_bottleneck}
\end{figure*}

\subsection{Photometric Bottleneck and Counterfactual Isolation}
To diagnose why coverage drops on challenging materials---and to understand how HPG foreground predictions (conditioned on frozen Depth Anything V2 features) limit instance recall---we trace every GT object in HouseCat6D under the uncapped production prompt budget (mean 61.89 prompts/image, vs.\ 58.16 matched prompt counts in the backbone comparison of Table~\ref{tab:app_hpg_depth_coverage}) through four stages: (1)~Prompt Hit (at least one prompt inside the GT mask), (2)~OLQA Top-1 Availability (at least one OLQA-selected candidate from any prompt in the image achieves $\mathrm{IoU}\ge 0.5$ with the GT mask; this image-wide availability is slightly higher than the prompt-assigned Object Recall@0.5 in the main paper, which restricts matching to prompts landing inside the target GT), (3)~After Foreground Filter (surviving candidate foreground overlap), and (4)~Final Output (Final Instance Recall@0.5, matched one-to-one at $\mathrm{IoU}\ge 0.5$ after GCSI). We group objects into the category proxies \emph{Glass} (transparent), \emph{Cutlery} (specular/thin), and \emph{Other} (remaining eight categories).

\begin{table}[!t]
\caption{Effect of HPG backbone features and relative depth on HouseCat6D under matched prompt counts per image (58.16 prompts/image on average).}
\label{tab:app_hpg_depth_coverage}
\centering
\footnotesize
\setlength{\tabcolsep}{3.5pt}
\renewcommand{\arraystretch}{1.04}
\begin{adjustbox}{max width=\linewidth}
\begin{tabular}{llccc}
\toprule
Category & HPG & Coverage (\%) $\uparrow$ & Mean Selected IoU $\uparrow$ & Object Recall (\%) $\uparrow$ \\
\midrule
\multirow{3}{*}{Glass}
 & DINOv2 & 87.661 & 0.8768 & 86.986 \\
 & DAV2 & 87.538 & 0.8845 & 86.618 \\
 & DAV2 + Relative Depth & \textbf{90.055} & \textbf{0.8847} & \textbf{88.950} \\
\midrule
\multirow{3}{*}{Cutlery}
 & DINOv2 & 67.187 & 0.7796 & 63.880 \\
 & DAV2 & 73.612 & \textbf{0.7841} & 70.930 \\
 & DAV2 + Relative Depth & \textbf{76.731} & 0.7720 & \textbf{72.863} \\
\midrule
\multirow{3}{*}{Other}
 & DINOv2 & 98.082 & 0.9049 & 97.855 \\
 & DAV2 & \textbf{99.184} & \textbf{0.9077} & \textbf{99.016} \\
 & DAV2 + Relative Depth & 99.074 & 0.9048 & 98.927 \\
\bottomrule
\end{tabular}
\end{adjustbox}
\end{table}

\noindent\textbf{HPG backbone features and relative depth.}
Table~\ref{tab:app_hpg_depth_coverage} compares HPG using frozen DINOv2, DAV2, and DAV2 with relative depth under matched prompt counts per image (the minimum prompt count across the three configurations, averaging 58.16 prompts/image). Compared with DINOv2, DAV2 improves pre-filtering Object Recall@0.5 on Cutlery from 63.88\% to 70.93\%. Adding relative depth further improves prompt Coverage by 2.52 and 3.12\,pp on Glass and Cutlery, respectively, with corresponding Object Recall gains of 2.33 and 1.93\,pp, while maintaining $\approx 99\%$ Object Recall on Other and comparable Mean Selected IoU. We therefore adopt DAV2 with relative depth in HPG.

As shown in Fig.~\ref{fig:app_hpg_bottleneck}(a), \emph{Other} objects exhibit minimal attrition across the entire pipeline (99.46\% prompt hit $\rightarrow$ 99.32\% Top-1 $\rightarrow$ 98.70\% after foreground filtering $\rightarrow$ 98.03\% Final Instance Recall@0.5). Crucially, for \emph{Glass} objects, HPG prompt coverage (\textbf{91.10\%}) and OLQA Top-1 candidate availability (\textbf{90.91\%}) are also high---demonstrating that HPG and frozen SAM already discover and segment most transparent instances. The dominant loss occurs specifically at the foreground overlap filter, where Glass retention drops from 90.91\% to \textbf{35.85\%} (Final Instance Recall@0.5: 35.67\%), and Cutlery drops from 76.92\% to \textbf{52.21\%} (Final Instance Recall@0.5: 49.97\%), because HPG foreground probabilities (conditioned on frozen DAV2 features) are lower over transparent and thin reflective surfaces.

Fig.~\ref{fig:app_hpg_bottleneck}(b) isolates the causal mechanism when relaxing $\tau_{\mathrm{fg}}$ from 0.90 to 0.80. Updating only the prompt coordinates while keeping the original foreground filter fixed changes Final Instance Recall@0.5 by merely $+0.10$\,pp overall ($+0.43$\,pp on Glass, $+0.06$\,pp on Cutlery). In contrast, relaxing only the foreground filter while keeping prompts fixed recovers \textbf{+2.84\,pp} overall (\textbf{+14.67\,pp} on Glass and \textbf{+11.67\,pp} on Cutlery) with only a $-0.06$\,pp shift in Overlap F, accounting for over 93\% of the joint recall gain ($+3.06$\,pp overall, $+14.79$\,pp Glass). Even with prompts fixed, relaxing only the foreground filter increases HouseCat6D FP by +2.00\,pp (19.97\% to 21.97\%, 95\% CI [+1.57, +2.44]; count MAE 1.68 to 1.98) and degrades OSD by $-0.39$\,pp Overlap F, $-0.52$\,pp Boundary F, $-1.19$\,pp Instance F, and +2.02\,pp FP. This finding directly confirms that adaptive or calibrated foreground filtering---rather than denser prompting---is the primary lever for further reducing missed transparent and thin reflective objects.

\noindent\textbf{Rationale for retaining the default operating point.}
Three considerations motivate keeping $(\tau_{\mathrm{fg}}=0.90,\ \tau_{\mathrm{fg,ov}}=0.75)$ as the default. First, \methodname\ uses one shared operating point across OCID, OSD, HouseCat6D, and grasping without target-domain tuning: lowering $\tau_{\mathrm{fg}}$ to 0.80 inflates FP and lowers Instance F on all three benchmarks while degrading OSD segmentation, and lowering $\tau_{\mathrm{fg,ov}}$ to 0.45 drops HouseCat6D Overlap F by $\approx 1.0$\,pp with +3.06--4.44\,pp higher FP (Fig.~\ref{fig:app_gcsi_pareto}). Second, suction grasping has asymmetric error costs: background false positives directly yield invalid grasps, whereas missing a secondary object allows picking another instance and re-scanning the scene after clearing. Accordingly, background inclusion causes 51 of 79 UOAIS-Net real-world failures (46 of 63 in Mixture) vs.\ zero for \methodname\ (Table~\ref{tab:real_failure_modes}), and stricter GCSI outperforms higher-recall NMS on SuctionNet-1Billion in Scene AP@50 (27.31\% vs.\ 26.25\%), Target AP@10 (18.27\% vs.\ 17.78\%), and Success@10 (61.64\% vs.\ 61.53\%) despite more missing targets (853 vs.\ 585; Table~\ref{tab:downstream_suction}). Third, high-recall deployments can lower $\tau_{\mathrm{fg}}$ (to 0.80) or $\tau_{\mathrm{fg,ov}}$ (to 0.45--0.60) directly at inference without retraining.

\begin{table}[!t]
\caption{Uncapped natural HPG prompt counts per image in the production pipeline ($\tau_{\mathrm{fg}}=0.90$, $\tau_h=0.012$).}
\label{tab:app_hpg_counts}
\centering
\footnotesize
\setlength{\tabcolsep}{4pt}
\renewcommand{\arraystretch}{1.05}
\begin{adjustbox}{max width=\linewidth}
\begin{tabular}{lrrrrrr}
\toprule
Dataset & Images & Mean & Median & P90 & Min--Max & Zero prompts \\
\midrule
OCID & 2,390 & 45.82 & 37 & 96 & 0--162 & 88 \\
OSD & 111 & 47.18 & 36 & 76 & 15--175 & 0 \\
HouseCat6D & 2,503 & 61.89 & 55 & 101 & 15--180 & 0 \\
\bottomrule
\end{tabular}
\end{adjustbox}
\end{table}

\begin{figure*}[!t]
\centering
\includegraphics[width=0.93\textwidth]{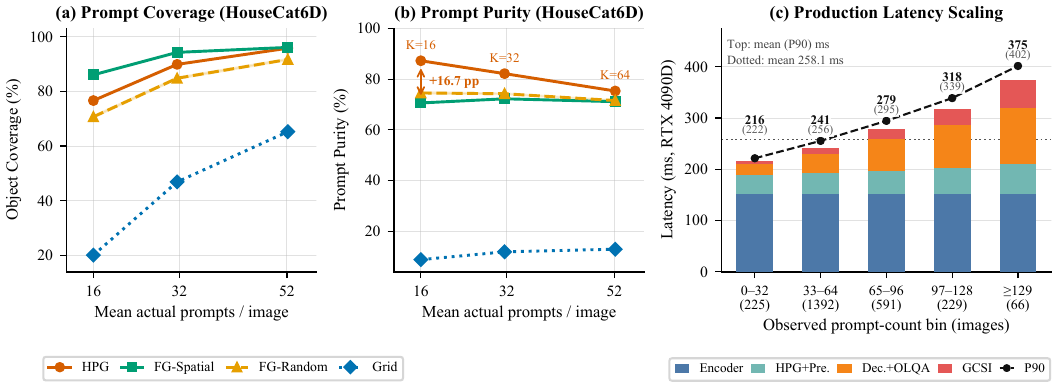}
\caption{Prompt budget scaling and production runtime scaling on HouseCat6D (2,503 images). (a)--(b)~Object Coverage at the prompt stage and Prompt Purity under fixed prompt budgets $K \in \{16, 32, 64\}$ (mean actual prompts: 16.0, 31.6, 52.2); shaded bands on FG-Random indicate $\pm 1$ SD over five sampling seeds. (c)~Measured production latency breakdown (RTX 4090D, FP32, batch size 1, $480\times 640$) grouped by bins of prompt count, with values above each bar showing mean and (P90) total latency in milliseconds.}
\label{fig:app_hpg_budget_latency}
\end{figure*}

\subsection{Natural Prompt Counts, Budget Saturation, and Runtime Scaling}
Table~\ref{tab:app_hpg_counts} summarizes the scene-adaptive prompt counts generated by HPG across all three benchmarks. HPG produces an average of 45.8--61.9 prompts per image (median 36--55, P90 76--101), which is $16\times$--$22\times$ sparser than the $32\times 32$ regular grid (1,024 prompts). The 88 frames with zero prompts in OCID correspond to empty tabletop scenes.

\noindent\textbf{Prompt budget saturation.}
Unlike the adaptive prompt evaluation in the main paper, Fig.~\ref{fig:app_hpg_budget_latency}(a)--(b) impose fixed prompt budgets $K \in \{16, 32, 64\}$ on HouseCat6D to compare prompting policies under matched point counts at the prompt stage (independent of downstream selector choice). Coverage rises steeply from $K=16$ to $K=32$ and approaches saturation by $K=64$ (95.77\% for HPG at an average of 52.2 actual prompts, closely matching the uncapped natural coverage of 96.59\% at 61.9 prompts). At every budget, HPG achieves the highest prompt purity, with its advantage over FG-Spatial widest under tight budgets: \textbf{+16.65\,pp} at $K=16$ (87.25\% vs.\ 70.60\%) and \textbf{+9.93\,pp} at $K=32$ (82.17\% vs.\ 72.24\%). The same pattern holds on OCID (HPG coverage $67.79\% \rightarrow 81.99\% \rightarrow 92.17\%$, purity 97.24--97.75\%) and OSD (coverage $78.27\% \rightarrow 87.34\% \rightarrow 93.46\%$, purity 94.60--96.06\%). Thus, a practical budget of 32--64 prompts captures nearly all discoverable objects while preserving high foreground purity.

\noindent\textbf{Measured runtime scaling with prompt count.}
Fig.~\ref{fig:app_hpg_budget_latency}(c) breaks down end-to-end production latency across all 2,503 HouseCat6D images on a single RTX 4090D GPU (FP32, batch size 1, $480\times 640$, 20 warmup frames; overall mean 258.14\,ms). Because SAM's image encoder runs once per frame, its cost remains constant at $\approx 151.8$\,ms across all prompt bins. The remaining stages grow smoothly with prompt count: from the 0--32 bin (mean 27.3 prompts) to the $\ge 129$ bin (mean 143.3 prompts), Decoder+OLQA scales from 21.9\,ms to 110.5\,ms, HPG+preprocessing from 36.5\,ms to 57.6\,ms, and GCSI+composition from 6.2\,ms to 54.4\,ms, yielding total mean latency from 216.5\,ms (P90 221.6\,ms) to 374.5\,ms (P90 401.8\,ms). Note that these bins reflect naturally varying scene clutter rather than synthetic prompt count interventions.

\section{HouseCat6D Official Partition Evaluation}
\label{app:housecat}

\begin{table*}[!t]
\caption{HouseCat6D evaluation broken down by official scene partition under the uniform 10-frame stride (\%). Official \textbf{Test} rows ($N=295$) are highlighted.}
\label{tab:app_housecat_splits}
\centering
\footnotesize
\setlength{\tabcolsep}{4.5pt}
\renewcommand{\arraystretch}{1.05}
\begin{adjustbox}{max width=\textwidth}
\begin{tabular}{llrrrrrrrrrrr}
\toprule
 & & & \multicolumn{3}{c}{Overlap} & \multicolumn{3}{c}{Boundary} & & \multicolumn{3}{c}{Instance (@$\mathrm{IoU}\ge 0.5$)} \\
\cmidrule(lr){4-6}\cmidrule(lr){7-9}\cmidrule(lr){11-13}
Architecture & Partition & $N$ & P $\uparrow$ & R $\uparrow$ & F $\uparrow$ & P $\uparrow$ & R $\uparrow$ & F $\uparrow$ & $\%75$ $\uparrow$ & IF $\uparrow$ & FP $\downarrow$ & Miss $\downarrow$ \\
\midrule
\multirow{4}{*}{SAM1 ViT-L}
 & \textbf{Test (official)} & \textbf{295} & \textbf{81.27} & \textbf{87.89} & \textbf{83.56} & \textbf{81.90} & \textbf{76.23} & \textbf{78.44} & \textbf{85.93} & \textbf{86.36} & \textbf{12.56} & \textbf{13.83} \\
 & Val & 141 & 73.52 & 91.22 & 79.08 & 76.44 & 86.81 & 80.36 & 91.21 & 85.40 & 18.67 & 8.58 \\
 & Train & 2,067 & 80.46 & 86.86 & 81.70 & 76.88 & 78.38 & 75.97 & 86.62 & 81.24 & 21.12 & 12.69 \\
 & All splits (main) & 2,503 & 80.16 & 87.22 & 81.77 & 77.44 & 78.60 & 76.51 & 86.80 & 82.08 & 19.97 & 12.60 \\
\midrule
\multirow{4}{*}{Semantic-SAM}
 & \textbf{Test (official)} & \textbf{295} & \textbf{80.35} & \textbf{86.63} & \textbf{82.49} & \textbf{80.06} & \textbf{73.02} & \textbf{75.76} & \textbf{83.32} & \textbf{85.17} & \textbf{12.86} & \textbf{15.76} \\
 & Val & 141 & 72.69 & 90.14 & 78.32 & 74.53 & 82.76 & 77.64 & 88.37 & 84.04 & 19.05 & 11.28 \\
 & Train & 2,067 & 79.01 & 85.90 & 80.44 & 76.34 & 74.93 & 73.97 & 83.90 & 79.90 & 21.85 & 14.70 \\
 & All splits (main) & 2,503 & 78.81 & 86.22 & 80.56 & 76.68 & 75.14 & 74.39 & 84.08 & 80.76 & 20.63 & 14.63 \\
\midrule
\multirow{4}{*}{SEEM Focal-T}
 & \textbf{Test (official)} & \textbf{295} & \textbf{78.26} & \textbf{80.05} & \textbf{78.18} & \textbf{68.67} & \textbf{61.11} & \textbf{64.31} & \textbf{79.12} & \textbf{80.74} & \textbf{19.87} & \textbf{17.59} \\
 & Val & 141 & 67.59 & 74.83 & 68.71 & 55.02 & 59.25 & 56.54 & 71.56 & 68.67 & 37.28 & 23.12 \\
 & Train & 2,067 & 76.73 & 78.99 & 75.89 & 64.23 & 61.21 & 61.41 & 77.52 & 74.52 & 28.94 & 18.64 \\
 & All splits (main) & 2,503 & 76.39 & 78.88 & 75.76 & 64.23 & 61.09 & 61.47 & 77.37 & 74.92 & 28.34 & 18.77 \\
\bottomrule
\end{tabular}
\end{adjustbox}
\end{table*}

In the main paper, because \methodname\ is trained solely on synthetic TOD scenes without any HouseCat6D training or threshold selection, we evaluate zero-shot transfer across all 2,503 images sampled every 10 frames from HouseCat6D. To facilitate direct comparison with future works following the official split protocol, Table~\ref{tab:app_housecat_splits} reports the complete breakdown across the disjoint official training ($N=2{,}067$), validation ($N=141$), and test ($N=295$) scene partitions using identical sampled image lists for all three frozen backbones.

Notably, across every evaluated architecture, the official sampled \textbf{Test} partition strictly exceeds the combined \emph{All splits} evaluation on all primary F-measures (Overlap F, Boundary F, and Instance F) as well as FP rate: SAM1 reaches \textbf{83.56\%} Overlap F, \textbf{78.44\%} Boundary F, and \textbf{86.36\%} Instance F on Test (vs.\ 81.77\%, 76.51\%, and 82.08\% on All splits, with FP dropping from 19.97\% to 12.56\%); Semantic-SAM improves from 80.56\% to \textbf{82.49\%} Overlap F; and SEEM improves from 75.76\% to \textbf{78.18\%} Overlap F. By contrast, $\%75$ and Miss Rate on Test are slightly less favorable than on All splits for SAM1 (85.93\% vs.\ 86.80\%; Miss 13.83\% vs.\ 12.60\%) and Semantic-SAM (83.32\% vs.\ 84.08\%; Miss 15.76\% vs.\ 14.63\%), while improving on SEEM. Thus, pooling all HouseCat6D partitions in the main evaluation provides a conservative---rather than inflated---estimate of zero-shot generalization on the primary F-measures. Consistent with the sampling protocol in the main paper, these Test metrics cover every tenth frame of the official test videos, providing a direct official split reference for future work.

\section{Cross-Architecture OLQA Feature Taps and Implementation Details}
\label{app:olqa}
\begin{table*}[!ht]
\caption{Cross-architecture \hdnetname\ feature taps, input construction, residual MLP specifications, and adapter parameter counts.}
\label{tab:app_olqa_specs}
\centering
\footnotesize
\setlength{\tabcolsep}{4.5pt}
\renewcommand{\arraystretch}{1.12}
\begin{adjustbox}{max width=\textwidth}
\begin{tabular}{llllllrr}
\toprule
Architecture & Frozen Backbone & Candidate Feature Tap & Normalization \& Fusion & Native Score $s_{ik}$ & Input Dim. & Residual MLP & Adapter Params \\
\midrule
SAM1 & ViT-L & Shared IoU token (256) + mask token (256) & Direct concat $[\mathbf{t}_{\mathrm{iou}}, \mathbf{t}_{\mathrm{mask}}^{(k)}]$ & Predicted mask IoU & 512 & $512{\rightarrow}256{\rightarrow}256{\rightarrow}1$ & 197,377 \\
Semantic-SAM & Swin-T & Final decoder layer IoU head input (256) & Direct decoder token $\mathbf{h}_{ik}$ & Native query quality & 256 & $256{\rightarrow}256{\rightarrow}256{\rightarrow}1$ & 131,841 \\
SEEM & Focal-T & Mask query $\mathbf{z}_{ik}$ (512) + spatial prompt $\mathbf{s}_i$ (512) & L2-norm $[\hat{\mathbf{z}}, \hat{\mathbf{s}}, \hat{\mathbf{z}}{\odot}\hat{\mathbf{s}}, s_{\mathrm{raw}}, s_{\mathrm{nat}}]$ & Softmax match score & 1,538 & $1538{\rightarrow}256{\rightarrow}256{\rightarrow}1$ & 460,033 \\
\bottomrule
\end{tabular}
\end{adjustbox}
\end{table*}

\begin{figure}[!t]
\centering
\includegraphics[width=\columnwidth]{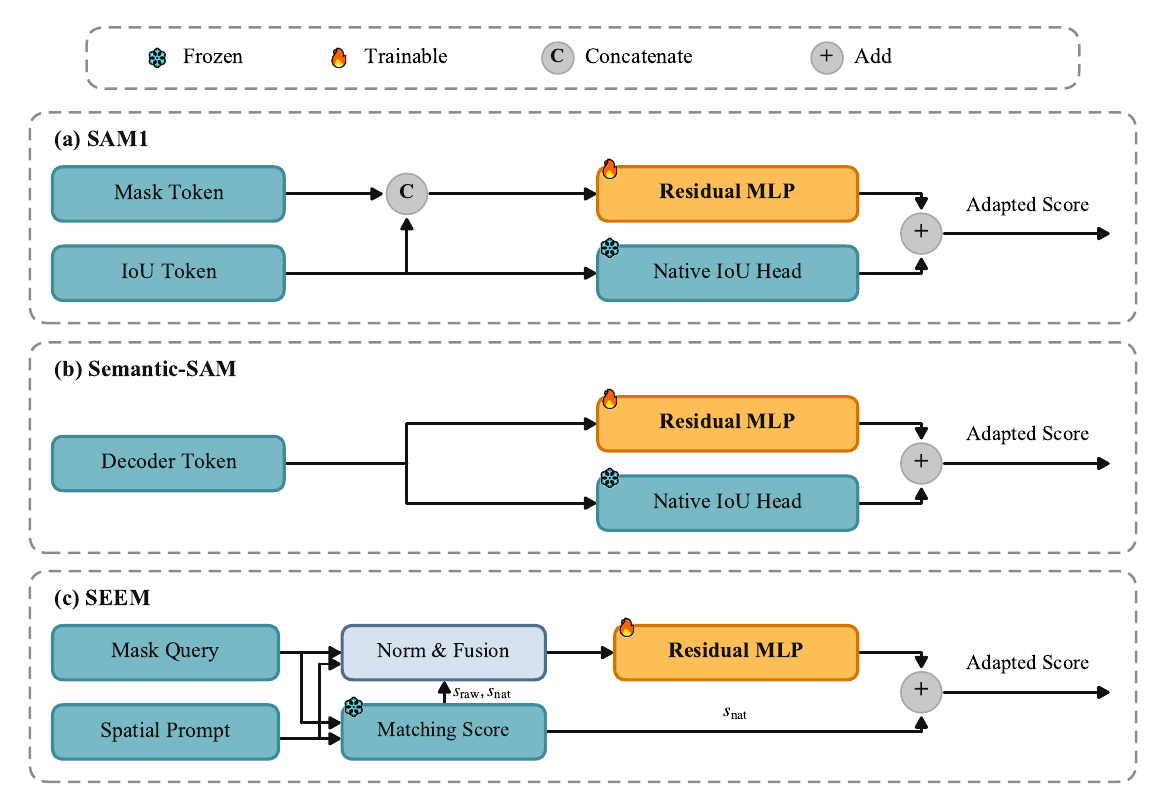}
\caption{Cross-architecture instantiation of \hdnetname\ across three frozen promptable foundation models: (a)~SAM1, (b)~Semantic-SAM, and (c)~SEEM. Each architecture taps its native decoder representations to feed an independent lightweight residual MLP added to its native candidate score, while keeping the pretrained mask generator frozen.}
\label{fig:app_olqa_architectures}
\end{figure}

Each promptable segmentation architecture uses an independently trained residual quality adapter $f_\theta(\mathbf{z}_{ik})$ with the same three-layer MLP structure (two 256-unit hidden layers with ReLU activations and a linear scalar residual output) and the same visible-object IoU supervision on 10\% of TOD scenes, while keeping the entire foundation mask generator frozen. No adapter weights are shared across architectures. Figure~\ref{fig:app_olqa_architectures} illustrates the three adapter designs, and Table~\ref{tab:app_olqa_specs} summarizes the exact feature taps and dimensionalities.

\noindent\textbf{SAM1.}
For each point prompt, SAM's mask decoder produces four internal mask tokens (one single-mask token and three multimask tokens) alongside a shared 256-dimensional IoU token. We concatenate the shared IoU token before each candidate's 256-dimensional mask token (forming a 512-dimensional input vector per candidate, observed shape $[1, 4, 512]$ per prompt) and retain the three multimask candidates during inference. The scalar residual is added directly to SAM's native IoU prediction. No additional feature normalization or external positional encoding is applied beyond the frozen decoder's internal encodings.

\noindent\textbf{Semantic-SAM.}
We tap the 256-dimensional candidate representation immediately entering the native IoU prediction head at the final decoder layer (cached shape $[N_p, 6, 256]$ across the six multi-granularity mask candidates per prompt) and add the learned MLP residual to the native candidate quality score, without extra normalization or positional encoding.

\noindent\textbf{SEEM.}
Unlike SAM and Semantic-SAM, SEEM scores candidate masks via cross-modal query--prompt matching rather than a calibrated IoU regression head. For each prompt, we extract the 512-dimensional spatial prompt embedding $\mathbf{s}_i$ and the 512-dimensional mask embeddings $\{\mathbf{z}_{ik}\}$ of the top-7 candidate shortlist, apply L2 normalization to obtain $\hat{\mathbf{z}}_{ik}$ and $\hat{\mathbf{s}}_i$, and form the 1,538-dimensional input $[\hat{\mathbf{z}}_{ik}, \hat{\mathbf{s}}_i, \hat{\mathbf{z}}_{ik} \odot \hat{\mathbf{s}}_i, s_{\mathrm{raw}}, s_{\mathrm{nat}}]$, where $s_{\mathrm{raw}}$ and $s_{\mathrm{nat}}$ denote the unnormalized and softmax-normalized prompt--mask matching scores, respectively. L2 normalization and the Hadamard interaction $\hat{\mathbf{z}}_{ik} \odot \hat{\mathbf{s}}_i$ explicitly expose cosine alignment between the prompt and candidate mask, enabling the residual MLP to map uncalibrated matching scores onto object-level IoU estimates.

\noindent\textbf{Model selection and release.}
All adapters are trained solely on synthetic TOD using validation loss for checkpoint selection; no checkpoint is tuned or selected using target dataset F-measures. Complete training/evaluation scripts, configuration files, and pretrained adapter checkpoints will be released publicly upon publication.

\section{Real-World Suction Evaluation Details}
\label{app:realworld}

\begin{figure*}[!t]
\centering
\includegraphics[width=\textwidth]{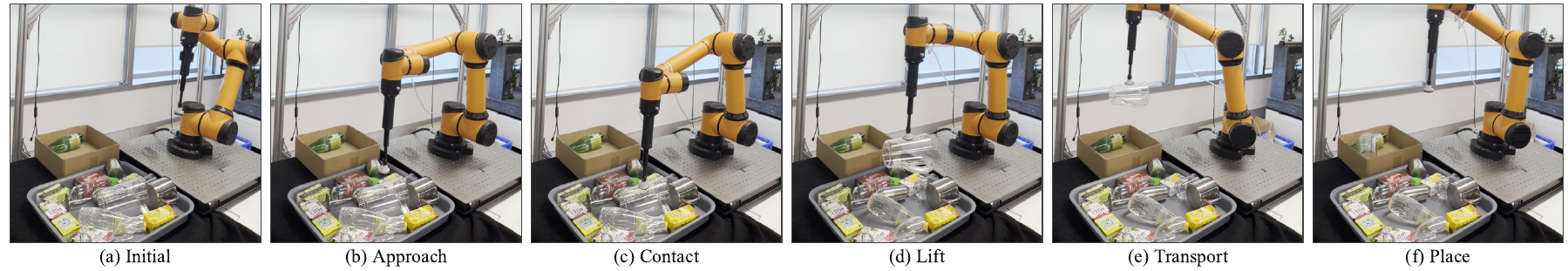}
\caption{Representative real-world suction grasping execution by \methodname\ in the \emph{Mixture} setting: the 6-DoF AUBO-i5 arm approaches, contacts, lifts, transports, and places a transparent pitcher from cluttered opaque, transparent, and reflective objects into the adjacent bin.}
\label{fig:app_realworld_execution}
\end{figure*}

\noindent\textbf{Hardware, sensing, and scene setup.}
The real-world evaluation in Section~\ref{sec:real_suction} uses a 6-DoF AUBO-i5 collaborative robot arm fitted with a pneumatic vacuum suction cup operating over a tabletop pick tray and an adjacent placement bin (Fig.~\ref{fig:app_realworld_execution}). An overhead Intel RealSense RGB camera ($1920\times 1080$ at 30\,Hz) and an overhead Photoneo PhoXi 3D Scanner are rigidly mounted above the workspace. Extrinsic RealSense--Photoneo registration projects 3D points into the RGB frame (with startup intrinsic verification), and Photoneo-to-robot hand-eye calibration maps planned suction poses into the AUBO-i5 base frame. Both segmentation methods operate on a fixed tabletop region of interest ($1100\times 700$ pixels, $[460, 330, 1560, 1030]$) before resizing to their respective network input sizes. The \emph{Opaque} setting uses 10 household objects, while \emph{Mixture} uses 15 opaque, transparent, and reflective objects; in \emph{Mixture}, UA-Depth (ViT-L encoder, 5 refinement iterations at $518\times 518$) restores missing or corrupted raw depth prior to suction prediction with UISN.

\begin{table}[!t]
\caption{Paired target filtering in the real-world suction evaluation ($245 \rightarrow 225$ pairs). Share is computed over all 245 initial pairs.}
\label{tab:app_realworld_filtering}
\centering
\footnotesize
\setlength{\tabcolsep}{3.5pt}
\renewcommand{\arraystretch}{1.05}
\begin{adjustbox}{max width=\linewidth}
\begin{tabular}{lrrrr}
\toprule
Category & Opaque & Mixture & Total & Share \\
\midrule
Initial paired targets & 116 & 129 & 245 & 100.0\% \\
\midrule
Excl.: workspace or execution-pose limit & $-2$ & $-2$ & $-4$ & 1.6\% \\
Excl.: depth quality check (both methods) & $-5$ & $0$ & $-5$ & 2.0\% \\
Excl.: depth quality check (UOAIS-Net only) & $-8$ & $0$ & $-8$ & 3.3\% \\
Excl.: depth quality check (\methodname\ only) & $-2$ & $0$ & $-2$ & 0.8\% \\
Excl.: baseline background selection & $0$ & $-1$ & $-1$ & 0.4\% \\
\midrule
\textbf{Retained paired evaluations (main paper)} & \textbf{99} & \textbf{126} & \textbf{225} & \textbf{91.8\%} \\
\bottomrule
\end{tabular}
\end{adjustbox}
\end{table}

\noindent\textbf{Excluded cases and trial filtering.}
Table~\ref{tab:app_realworld_filtering} accounts for all 20 excluded pairs (8.2\%) between the 245 initial paired targets (116 Opaque, 129 Mixture) and the 225 retained paired evaluations (99 Opaque, 126 Mixture) reported in the main paper:
(1)~\emph{Workspace or execution-pose limits} (4 pairs: 2 Opaque, 2 Mixture; 1.6\%), where robot kinematics or workspace safety bounds rejected execution independently of segmentation;
(2)~\emph{Depth quality check rejection} (15 pairs, all in Opaque using raw Photoneo depth; 6.1\%), where the planner's geometric checks rejected a mask due to missing depth at the mask center, excessive missing depth within the mask, or a local surface discontinuity (5 on both methods from raw depth holes on the object, 8 on UOAIS-Net only---5 caused by merging adjacent objects across an inter-object gap plus 3 depth quality check rejections---and 2 on \methodname\ only); and
(3)~\emph{Baseline background selection} (1 Mixture pair; 0.4\%), where UOAIS-Net associated the sampled target prompt with a background region while \methodname\ segmented the object.

{\small
    \bibliographystyle{IEEEtran}
    \bibliography{reference}

@String(CVPR= {IEEE Conf. Comput. Vis. Pattern Recog.})

@String(ICCV= {Int. Conf. Comput. Vis.})

@String(ECCV= {Eur. Conf. Comput. Vis.})

@String(CVPR  = {CVPR})

@String(ICCV  = {ICCV})

@String(ECCV  = {ECCV})

@inproceedings{kirillov2023segment,
  title={Segment anything},
  author={Kirillov, Alexander and Mintun, Eric and Ravi, Nikhila and Mao, Hanzi and Rolland, Chloe and Gustafson, Laura and Xiao, Tete and Whitehead, Spencer and Berg, Alexander C and Lo, Wan-Yen and others},
  booktitle={Proceedings of the IEEE/CVF International Conference on Computer Vision},
  pages={4015--4026},
  year={2023}
}

@article{vaswani2017attention,
  title={Attention is all you need},
  author={Vaswani, Ashish and Shazeer, Noam and Parmar, Niki and Uszkoreit, Jakob and Jones, Llion and Gomez, Aidan N and Kaiser, {\L}ukasz and Polosukhin, Illia},
  journal={Advances in neural information processing systems},
  volume={30},
  year={2017}
}

@inproceedings{xiang2020learning,
    Author = {Yu Xiang and Christopher Xie and Arsalan Mousavian and Dieter Fox},
    Title = {Learning RGB-D Feature Embeddings for Unseen Object Instance Segmentation},
    booktitle = {Conference on Robot Learning (CoRL)},
    Year = {2020}
}

@inproceedings{cheng2022masked,
  title={Masked-attention mask transformer for universal image segmentation},
  author={Cheng, Bowen and Misra, Ishan and Schwing, Alexander G and Kirillov, Alexander and Girdhar, Rohit},
  booktitle={Proceedings of the IEEE/CVF Conference on Computer Vision and Pattern Recognition},
  pages={1290--1299},
  year={2022}
}

@article{loshchilov2017decoupled,
  title={Decoupled weight decay regularization},
  author={Loshchilov, Ilya and Hutter, Frank},
  journal={arXiv preprint arXiv:1711.05101},
  year={2017}
}

@inproceedings{suchi2019easylabel,
  title={EasyLabel: A semi-automatic pixel-wise object annotation tool for creating robotic RGB-D datasets},
  author={Suchi, Markus and Patten, Timothy and Fischinger, David and Vincze, Markus},
  booktitle={2019 International Conference on Robotics and Automation (ICRA)},
  pages={6678--6684},
  year={2019},
  organization={IEEE}
}

@inproceedings{richtsfeld2012segmentation,
  title={Segmentation of unknown objects in indoor environments},
  author={Richtsfeld, Andreas and M{\"o}rwald, Thomas and Prankl, Johann and Zillich, Michael and Vincze, Markus},
  booktitle={2012 IEEE/RSJ International Conference on Intelligent Robots and Systems},
  pages={4791--4796},
  year={2012},
  organization={IEEE}
}

@inproceedings{mousavian20196,
  title={6-dof graspnet: Variational grasp generation for object manipulation},
  author={Mousavian, Arsalan and Eppner, Clemens and Fox, Dieter},
  booktitle={Proceedings of the IEEE/CVF International Conference on Computer Vision},
  pages={2901--2910},
  year={2019}
}

@inproceedings{xie2020best,
  title={The best of both modes: Separately leveraging rgb and depth for unseen object instance segmentation},
  author={Xie, Christopher and Xiang, Yu and Mousavian, Arsalan and Fox, Dieter},
  booktitle={Conference on robot learning},
  pages={1369--1378},
  year={2020},
  organization={PMLR}
}

@article{ochs2013segmentation,
  title={Segmentation of moving objects by long term video analysis},
  author={Ochs, Peter and Malik, Jitendra and Brox, Thomas},
  journal={IEEE transactions on pattern analysis and machine intelligence},
  volume={36},
  number={6},
  pages={1187--1200},
  year={2013},
  publisher={IEEE}
}

@article{xie2021unseen,
  title={Unseen object instance segmentation for robotic environments},
  author={Xie, Christopher and Xiang, Yu and Mousavian, Arsalan and Fox, Dieter},
  journal={IEEE Transactions on Robotics},
  volume={37},
  number={5},
  pages={1343--1359},
  year={2021},
  publisher={IEEE}
}

@inproceedings{back2022unseen,
  title={Unseen object amodal instance segmentation via hierarchical occlusion modeling},
  author={Back, Seunghyeok and Lee, Joosoon and Kim, Taewon and Noh, Sangjun and Kang, Raeyoung and Bak, Seongho and Lee, Kyoobin},
  booktitle={2022 International Conference on Robotics and Automation (ICRA)},
  pages={5085--5092},
  year={2022},
  organization={IEEE}
}

@inproceedings{he2017mask,
  title={Mask r-cnn},
  author={He, Kaiming and Gkioxari, Georgia and Doll{\'a}r, Piotr and Girshick, Ross},
  booktitle={Proceedings of the IEEE international conference on computer vision},
  pages={2961--2969},
  year={2017}
}

@inproceedings{danielczuk2019segmenting,
  title={Segmenting unknown 3d objects from real depth images using mask r-cnn trained on synthetic data},
  author={Danielczuk, Michael and Matl, Matthew and Gupta, Saurabh and Li, Andrew and Lee, Andrew and Mahler, Jeffrey and Goldberg, Ken},
  booktitle={2019 International Conference on Robotics and Automation (ICRA)},
  pages={7283--7290},
  year={2019},
  organization={IEEE}
}

@inproceedings{durner2021unknown,
  title={Unknown object segmentation from stereo images},
  author={Durner, Maximilian and Boerdijk, Wout and Sundermeyer, Martin and Friedl, Werner and M{\'a}rton, Zolt{\'a}n-Csaba and Triebel, Rudolph},
  booktitle={2021 IEEE/RSJ International Conference on Intelligent Robots and Systems (IROS)},
  pages={4823--4830},
  year={2021},
  organization={IEEE}
}

@inproceedings{xie2022rice,
  title={Rice: Refining instance masks in cluttered environments with graph neural networks},
  author={Xie, Chris and Mousavian, Arsalan and Xiang, Yu and Fox, Dieter},
  booktitle={Conference on Robot Learning},
  pages={1655--1665},
  year={2022},
  organization={PMLR}
}

@inproceedings{lu2022mean,
  author={Lu, Yangxiao and Chen, Yuqiao and Ruozzi, Nicholas and Xiang, Yu},
  booktitle={2024 IEEE International Conference on Robotics and Automation (ICRA)}, 
  title={Mean Shift Mask Transformer for Unseen Object Instance Segmentation}, 
  year={2024},
  volume={},
  number={},
  pages={2760-2766},
  doi={10.1109/ICRA57147.2024.10610943}}

@book{gonzales1987digital,
  title={Digital image processing},
  author={Gonzales, Rafael C and Wintz, Paul},
  year={1987},
  publisher={Addison-Wesley Longman Publishing Co., Inc.}
}

@inproceedings{zhou2019objects,
  title={Objects as Points},
  author={Zhou, Xingyi and Wang, Dequan and Kr{\"a}henb{\"u}hl, Philipp},
  booktitle={arXiv preprint arXiv:1904.07850},
  year={2019}
}

@inproceedings{jung2024housecat6d,
  title={HouseCat6D-A Large-Scale Multi-Modal Category Level 6D Object Perception Dataset with Household Objects in Realistic Scenarios},
  author={Jung, HyunJun and Wu, Shun-Cheng and Ruhkamp, Patrick and Zhai, Guangyao and Schieber, Hannah and Rizzoli, Giulia and Wang, Pengyuan and Zhao, Hongcheng and Garattoni, Lorenzo and Meier, Sven and others},
  booktitle={Proceedings of the IEEE/CVF Conference on Computer Vision and Pattern Recognition},
  pages={22498--22508},
  year={2024}
}

@article{wu2020cgnet,
  title={Cgnet: A light-weight context guided network for semantic segmentation},
  author={Wu, Tianyi and Tang, Sheng and Zhang, Rui and Cao, Juan and Zhang, Yongdong},
  journal={IEEE Transactions on Image Processing},
  volume={30},
  pages={1169--1179},
  year={2020},
  publisher={IEEE}
}

@inproceedings{nguyen2024gigapose,
  title={Gigapose: Fast and robust novel object pose estimation via one correspondence},
  author={Nguyen, Van Nguyen and Groueix, Thibault and Salzmann, Mathieu and Lepetit, Vincent},
  booktitle={Proceedings of the IEEE/CVF Conference on Computer Vision and Pattern Recognition},
  pages={9903--9913},
  year={2024}
}

@inproceedings{labbe2022megapose,
  title={MegaPose: 6D Pose Estimation of Novel Objects via Render \& Compare},
  author={Labb{\'e}, Yann and Manuelli, Lucas and Mousavian, Arsalan and Tyree, Stephen and Birchfield, Stan and Tremblay, Jonathan and Carpentier, Justin and Aubry, Mathieu and Fox, Dieter and Sivic, Josef},
  booktitle={CoRL 2022-Conference on Robot Learning},
  year={2022}
}

@inproceedings{wen2022transgrasp,
  title={Transgrasp: Grasp pose estimation of a category of objects by transferring grasps from only one labeled instance},
  author={Wen, Hongtao and Yan, Jianhang and Peng, Wanli and Sun, Yi},
  booktitle={European Conference on Computer Vision},
  pages={445--461},
  year={2022},
  organization={Springer}
}

@article{cao2024uncertainty,
  title={Uncertainty-Aware Suction Grasping for Cluttered Scenes},
  author={Cao, Rui and Yang, Biqi and Li, Yichuan and Fu, Chi-Wing and Heng, Pheng-Ann and Liu, Yun-Hui},
  journal={IEEE Robotics and Automation Letters},
  year={2024},
  publisher={IEEE}
}

@article{suctionnet,
  title={{SuctionNet-1Billion}: A Large-Scale Benchmark for Suction Grasping},
  author={Cao, Hanwen and Fang, Hao-Shu and Liu, Wenhai and Lu, Cewu},
  journal={IEEE Robotics and Automation Letters},
  year={2021},
  volume={6},
  number={4},
  pages={8718--8725},
  doi={10.1109/LRA.2021.3115406}
}

@inproceedings{manuelli2019kpam,
  title={kpam: Keypoint affordances for category-level robotic manipulation},
  author={Manuelli, Lucas and Gao, Wei and Florence, Peter and Tedrake, Russ},
  booktitle={The International Symposium of Robotics Research},
  pages={132--157},
  year={2019},
  organization={Springer}
}

@inproceedings{chen2023learning,
  title={Learning to grasp clothing structural regions for garment manipulation tasks},
  author={Chen, Wei and Lee, Dongmyoung and Chappell, Digby and Rojas, Nicolas},
  booktitle={2023 IEEE/RSJ International Conference on Intelligent Robots and Systems (IROS)},
  pages={4889--4895},
  year={2023},
  organization={IEEE}
}

@INPROCEEDINGS{2024EquivAct,
  author={Yang, Jingyun and Deng, Congyue and Wu, Jimmy and Antonova, Rika and Guibas, Leonidas and Bohg, Jeannette},
  booktitle={2024 IEEE International Conference on Robotics and Automation (ICRA)}, 
  title={EquivAct: SIM(3)-Equivariant Visuomotor Policies beyond Rigid Object Manipulation}, 
  year={2024},
  volume={},
  number={},
  pages={9249-9255},
  doi={10.1109/ICRA57147.2024.10611491}}

@inproceedings{wang2020solo,
  title     =  {{SOLO}: Segmenting Objects by Locations},
  author    =  {Wang, Xinlong and Kong, Tao and Shen, Chunhua and Jiang, Yuning and Li, Lei},
  booktitle =  {Proc. Eur. Conf. Computer Vision (ECCV)},
  year      =  {2020}
}

@inproceedings{yang2023improving,
  title={On Improving Boundary Quality of Instance Segmentation in Cluttered and Chaotic Scenarios},
  author={Yang, Biqi and Gao, Xiaojie and Li, Xianzhi and Liu, Yun-Hui and Fu, Chi-Wing and Heng, Pheng-Ann},
  booktitle={2023 IEEE International Conference on Robotics and Automation (ICRA)},
  pages={9310--9316},
  year={2023},
  organization={IEEE}
}

@inproceedings{lin2024sam,
  title={Sam-6d: Segment anything model meets zero-shot 6d object pose estimation},
  author={Lin, Jiehong and Liu, Lihua and Lu, Dekun and Jia, Kui},
  booktitle={Proceedings of the IEEE/CVF Conference on Computer Vision and Pattern Recognition},
  pages={27906--27916},
  year={2024}
}

@inproceedings{hulora,
  title={LoRA: Low-Rank Adaptation of Large Language Models},
  author={Hu, Edward J and Wallis, Phillip and Allen-Zhu, Zeyuan and Li, Yuanzhi and Wang, Shean and Wang, Lu and Chen, Weizhu and others},
  booktitle={International Conference on Learning Representations},
  year={2022}
}

@article{li2022sim,
  title={A sim-to-real object recognition and localization framework for industrial robotic bin picking},
  author={Li, Xianzhi and Cao, Rui and Feng, Yidan and Chen, Kai and Yang, Biqi and Fu, Chi-Wing and Li, Yichuan and Dou, Qi and Liu, Yun-Hui and Heng, Pheng-Ann},
  journal={IEEE Robotics and Automation Letters},
  volume={7},
  number={2},
  pages={3961--3968},
  year={2022},
  publisher={IEEE}
}

@article{zisvfm,
  title={Zisvfm: Zero-shot object instance segmentation in indoor robotic environments with vision foundation models},
  author={Zhang, Ying and Yin, Maoliang and Bi, Wenfu and Yan, Haibao and Bian, Shaohan and Zhang, Cui-Hua and Hua, Changchun},
  journal={IEEE Transactions on Robotics},
  volume={41},
  pages={1568--1580},
  year={2025},
  publisher={IEEE}
}

@article{depth_anything_v2,
  title={Depth Anything V2},
  author={Yang, Lihe and Kang, Bingyi and Huang, Zilong and Zhao, Zhen and Xu, Xiaogang and Feng, Jiashi and Zhao, Hengshuang},
  journal={arXiv:2406.09414},
  year={2024}
}

@inproceedings{ranftl2021vision,
  title={Vision transformers for dense prediction},
  author={Ranftl, Ren{\'e} and Bochkovskiy, Alexey and Koltun, Vladlen},
  booktitle={2021 IEEE/CVF International Conference on Computer Vision (ICCV)},
  pages={12159--12168},
  year={2021},
  organization={IEEE}
}

@article{ravi2024sam2,
  title={{SAM 2}: Segment Anything in Images and Videos},
  author={Ravi, Nikhila and Gabeur, Valentin and Hu, Yuan-Ting and Hu, Ronghang and Ryali, Chaitanya and Ma, Tengyu and others},
  journal={arXiv preprint arXiv:2408.00714},
  year={2024}
}

@article{fu2024taylor,
  title={Taylor Neural Network for Unseen Object Instance Segmentation in Hierarchical Grasping},
  author={Fu, K. and Dang, X. and Zhang, Y.},
  journal={IEEE/ASME Transactions on Mechatronics},
  volume={29},
  number={5},
  pages={3485--3496},
  year={2024}
}

@INPROCEEDINGS{sam_adapter,
  author={Chen, Tianrun and Zhu, Lanyun and Ding, Chaotao and Cao, Runlong and Wang, Yan and Zhang, Shangzhan and Li, Zejian and Sun, Lingyun and Zang, Ying and Mao, Papa},
  booktitle={2023 IEEE/CVF International Conference on Computer Vision Workshops (ICCVW)}, 
  title={SAM-Adapter: Adapting Segment Anything in Underperformed Scenes}, 
  year={2023},
  volume={},
  number={},
  pages={3359-3367},
  doi={10.1109/ICCVW60793.2023.00361}
}

@inproceedings{wang2022freesolo,
  title={Freesolo: Learning to segment objects without annotations},
  author={Wang, Xinlong and Yu, Zhiding and De Mello, Shalini and Kautz, Jan and Anandkumar, Anima and Shen, Chunhua and Alvarez, Jose M},
  booktitle={2022 IEEE/CVF Conference on Computer Vision and Pattern Recognition (CVPR)},
  pages={14156--14166},
  year={2022},
  organization={IEEE}
}

@inproceedings{wang2023cut,
  title={Cut and learn for unsupervised object detection and instance segmentation},
  author={Wang, Xudong and Girdhar, Rohit and Yu, Stella X and Misra, Ishan},
  booktitle={Proceedings of the IEEE/CVF conference on computer vision and pattern recognition},
  pages={3124--3134},
  year={2023}
}

@inproceedings{li2024segment,
  title={Segment and recognize anything at any granularity},
  author={Li, Feng and Zhang, Hao and Sun, Peize and Zou, Xueyan and Liu, Shilong and Li, Chunyuan and Yang, Jianwei and Zhang, Lei and Gao, Jianfeng},
  booktitle={European Conference on Computer Vision},
  pages={467--484},
  year={2024},
  organization={Springer}
}

@article{zou2023segment,
  title={Segment everything everywhere all at once},
  author={Zou, Xueyan and Yang, Jianwei and Zhang, Hao and Li, Feng and Li, Linjie and Wang, Jianfeng and Wang, Lijuan and Gao, Jianfeng and Lee, Yong Jae},
  journal={Advances in neural information processing systems},
  volume={36},
  pages={19769--19782},
  year={2023}
}

@inproceedings{carion2026sam3,
  title={{SAM 3}: Segment Anything with Concepts},
  author={Carion, Nicolas and Gustafson, Laura and Hu, Yuan-Ting and Debnath, Shoubhik and Hu, Ronghang and Suris Coll-Vinent, Didac and Ryali, Chaitanya and Vasudev Alwala, Kalyan and Khedr, Haitham and Huang, Andrew and others},
  booktitle={International Conference on Learning Representations},
  year={2026},
  url={https://arxiv.org/abs/2511.16719}
}

@article{cao2026uadepth,
  title={{UA-Depth}: Uncertainty-Aware Depth Restoration for Material-Agnostic Object Manipulation},
  author={Cao, Rui and Pan, Jia-Hui and Zhang, Yin and Meng, Qiwei and Li, Jiapeng and Wang, Gang and Zhu, Runsong and Chen, Wei and Fu, Chi-Wing and Wang, Jiangliu and Liu, Yun-Hui},
  journal={arXiv preprint arXiv:2611.01645},
  year={2026},
  url={https://arxiv.org/abs/2611.01645}
}

@inproceedings{huang2019mask,
  title={Mask Scoring {R-CNN}},
  author={Huang, Zhaojin and Huang, Lichao and Gong, Yongchao and Huang, Chang and Wang, Xinggang},
  booktitle={Proceedings of the IEEE/CVF Conference on Computer Vision and Pattern Recognition (CVPR)},
  pages={6409--6418},
  year={2019},
  doi={10.1109/CVPR.2019.00657}
}

@inproceedings{fang2024embodied,
  title={Embodied Uncertainty-Aware Object Segmentation},
  author={Fang, Xiaolin and Kaelbling, Leslie Pack and Lozano-P{\'e}rez, Tom{\'a}s},
  booktitle={2024 IEEE/RSJ International Conference on Intelligent Robots and Systems (IROS)},
  pages={2639--2646},
  year={2024},
  doi={10.1109/IROS58592.2024.10801562},
  url={https://arxiv.org/abs/2408.04760}
}
}
\end{document}